\documentclass[sigconf, nonacm]{acmart}

\usepackage{pvldb}

\usepackage{amsfonts}       
\usepackage{amsmath}
\usepackage{amssymb}
\usepackage{nicefrac} 
\usepackage[ruled, vlined, linesnumbered]{algorithm2e}
\usepackage{enumerate}
\usepackage{enumitem}
\usepackage{geometry}
\usepackage{tikz}
\usetikzlibrary{shapes.geometric}
\usepackage{arydshln}
\usepackage{xspace}
\usepackage{amssymb}
\usepackage{bbding}
\usepackage{tcolorbox}
\usepackage{multirow}
\usepackage{threeparttable}
\usepackage{soul}
\usepackage{color}

\usepackage[capitalize,noabbrev]{cleveref}
\usepackage{makecell}
\usepackage{wrapfig}
\usepackage{caption}
\usepackage{subcaption}
\usepackage{pifont}
\usepackage{tikz}
\usepackage{hyperref}
\usepackage[thicklines]{cancel}
\SetKwInput{KwInput}{Input} 
\SetKwInput{KwOutput}{Output}  
\usepackage{graphicx} 
\usepackage[export]{adjustbox}
\usepackage{balance}
\usepackage[table]{xcolor}
\definecolor{lightpurple}{HTML}{c6c6ff}

\setulcolor{red}

\def\header{\vspace{2mm} \noindent}

\renewcommand\vldbdoi{XX.XX/XXX.XX}
\renewcommand\vldbpages{XXX-XXX}
\renewcommand\vldbavailabilityurl{https://github.com/Lucas-PJ/CascadeBench} 

\begin{document}
\title{From Leakage to Fidelity: Reliable Benchmarking for \\Temporal Cascade Prediction}

\author{Jie Peng}
\affiliation{%
  \institution{Renmin University of China}
  \city{Beijing}
  \state{Beijing}
  \country{China}
}
\email{peng\_jie@ruc.edu.cn}

\author{Rui	Wang}
\affiliation{
  \institution{Alibaba}
  \city{Beijing}
  \state{Beijing}
  \country{China}
}
\email{ruiwang0630@gmail.com}

\author{Qiang Wang}
\affiliation{
  \institution{Alibaba}
  \city{Beijing}
  \state{Beijing}
  \country{China}
}
\email{feifan.wq@taobao.com}

\author{Zhewei Wei}
\affiliation{
  \institution{Renmin University of China}
  \city{Beijing}
  \state{Beijing}
  \country{China}
}
\email{zhewei@ruc.edu.cn}

\author{{Bin Tong}}
\affiliation{
  \institution{Alibaba}
  \city{Beijing}
  \state{Beijing}
  \country{China}
}
\email{tongbin.tb@alibaba-inc.com}

\author{Guan Wang}
\affiliation{
  \institution{Alibaba}
  \city{Beijing}
  \state{Beijing}
  \country{China}
}
\email{shangfeng.wg@taobao.com}

\author{Bo Zheng}
\affiliation{
  \institution{Alibaba}
  \city{Beijing}
  \state{Beijing}
  \country{China}
}
\email{bozheng@alibaba-inc.com}

\begin{abstract}
Temporal cascade prediction is widely studied, yet its empirical foundations remain fragile. Most existing works report results under random cascade splits that mix past and future signals, rely on datasets with limited features and no downstream conversion labels, and compare increasingly complex models without systematically examining whether benchmark conclusions are protocol-dependent. This paper argues that the field should move from leakage-prone evaluation toward fidelity-aware benchmarking. We introduce a protocol suite and renewed evaluation standard for temporal cascade prediction, centered on the Full Temporal protocol, overlap-based leakage diagnostics, and analyses of performance inflation and temporal drift. To broaden the scope of benchmark tasks, we also present Taoke, a real-world e-commerce cascade dataset with rich promoter/product features and observed purchase conversions, enabling both first-stage popularity forecasting and second-stage conversion forecasting under a shared benchmark asset. Finally, we include CasTemp as a lightweight reference method and additionally probe a larger same-task internal extension to verify that this pipeline remains operational at substantially greater scale. Together, these components turn cascade prediction from a protocol-sensitive leaderboard exercise into a more reliable analysis and benchmarking problem, while still providing a practical reference pipeline for large-scale evaluation and conversion-aware modeling.
\end{abstract}

\maketitle

\vldbtopmatter

\section{Introduction}\label{intro}

Temporal cascade prediction studies how the popularity of an information item evolves after an initial observation period.
The problem appears in social media reposting \cite{weng2013viralitytwitter,cheng2014can}, citation accumulation \cite{min2017quantifying}, traffic flow distribution \cite{han2024bigst}, and e-commerce promotion \cite{liu2016influence}, where practitioners care not only about future diffusion volume but also about downstream business outcomes such as user engagement or purchases.
Forecasting the propagation scale (i.e., popularity) of these cascades not only helps researchers better understand the mechanisms of information spread but is also crucial for numerous applications, including detecting viral misinformation \cite{leskovec2007dynamics,weng2013viralitytwitter}, optimizing social media marketing strategies \cite{dave2011modelling,zhang2025sagraph}, and enhancing e-commerce recommendation systems \cite{gao2019learning,petrov2025efficient,peng2026can}.
Because of this broad relevance, the area has produced a long line of increasingly sophisticated models, ranging from feature-based regressors and recurrent architectures to graph neural networks, dynamic-memory models, and diffusion-style generators \cite{szabo2010predicting,cao2017deephawkes,chen2019information,xu2021casflow,Lu2023ContinuousTimeGL,cheng2024information}.
However, we argue that the central bottleneck of the field is no longer another modeling module, but the reliability of the benchmark itself.
Three issues motivate this view.

\header{\textbf{Protocol validity is under-examined.}}
Most prior work evaluates models with random cascade-level splits \cite{xu2021casflow,Lu2023ContinuousTimeGL,cheng2024information}.
This practice is convenient, but it weakens the temporal semantics of forecasting: training and test cascades can share future-era activity patterns, overlapping users, and globally synchronized bursts.
As a result, a model may benefit from shortcut signals that would not be available in a real deployment setting.
If benchmark conclusions depend strongly on such optimistic splits, then leaderboard gains do not necessarily translate into reliable forecasting ability.

\header{\textbf{Current datasets under-support analysis and practical tasks.}}
Widely used public datasets typically expose only propagation structures and timestamps, with limited cascade attributes, limited user features, and no downstream conversion labels.
This makes it hard to study why models succeed or fail, and it also narrows the task definition to first-stage popularity only.
In many real systems, though, the relevant question is not merely whether a cascade spreads, but whether it leads to actions such as clicks, comments, or purchases \cite{liu2007arsa,bodapati2008recommendation,liu2010s,thelwall2018social}.
Benchmarking should therefore cover richer data and richer targets.

\header{\textbf{Model complexity has outpaced benchmark scrutiny.}}
Recent methods introduce increasingly heavy machinery to capture structural and temporal dependencies.
Yet if the evaluation protocol itself is optimistic, it becomes unclear whether this complexity is learning genuine diffusion dynamics or simply exploiting artifacts in the split.
For an analysis-and-benchmark paper, an important question is therefore not only ``\emph{which model wins},'' but also ``\emph{which conclusions remain stable once the protocol becomes realistic}.''

\begin{figure}[t]
\centering
\vskip 0.1in
\resizebox{\columnwidth}{!}{
\begin{tikzpicture}[font=\footnotesize]
    \fill[gray!6, rounded corners=2pt] (0,0) rectangle (7.8,4.35);
    \draw[gray!35, rounded corners=2pt] (0,0) rectangle (7.8,4.35);

    \node[anchor=west, font=\bfseries\footnotesize] at (0.1,4.10) {Protocol correction changes ranks, leakage, and apparent difficulty.};

    \draw[gray!35, rounded corners=2pt, fill=white] (0.2,0.25) rectangle (3.4,3.85);
    
    \node[anchor=west, font=\bfseries\scriptsize] at (0.35,3.65) {Rank Changes on APS};
    
    \node[font=\scriptsize] at (0.94,3.23) {Random};
    \node[font=\scriptsize] at (0.94,3.00) {Cascade};
    \node[font=\scriptsize] at (1.86,3.23) {Full};
    \node[font=\scriptsize] at (1.86,3.00) {Temporal};
    \node[font=\scriptsize] at (2.78,3.23) {Inductive};
    \node[font=\scriptsize] at (2.78,3.00) {User};
    
    \definecolor{crownyellow}{RGB}{255,215,0}
    \definecolor{crowngold}{RGB}{218,165,32}
    \fill[crownyellow, draw=crowngold, line width=0.7pt, line join=round]
        (0.23,2.74) --          
        (0.23,2.80) --          
        (0.27,2.88) --          
        (0.31,2.80) --          
        (0.37,2.92) --          
        (0.43,2.80) --          
        (0.47,2.88) --          
        (0.51,2.80) --          
        (0.51,2.74) --          
        cycle;
    
    \node[anchor=east, font=\scriptsize] at (0.52,2.55) {1};
    \node[anchor=east, font=\scriptsize] at (0.52,1.70) {2};
    \node[anchor=east, font=\scriptsize] at (0.52,0.85) {3};

    \fill[blue!50] (0.54,2.34) rectangle (1.34,2.76);
    \fill[blue!30] (0.54,1.49) rectangle (1.34,1.91);
    \fill[blue!12] (0.54,0.64) rectangle (1.34,1.06);
    \node[font=\scriptsize] at (0.94,2.55) {CasFlow};
    \node[font=\scriptsize] at (0.94,1.70) {CasTemp};
    \node[font=\scriptsize] at (0.94,0.85) {CasCN};
    
    \fill[blue!50] (1.46,2.34) rectangle (2.26,2.76);
    \fill[blue!30] (1.46,1.49) rectangle (2.26,1.91);
    \fill[blue!12] (1.46,0.64) rectangle (2.26,1.06);
    \node[font=\scriptsize] at (1.86,2.55) {CasTemp};
    \node[font=\scriptsize] at (1.86,1.70) {VNOIP};
    \node[font=\scriptsize] at (1.86,0.85) {CasCN};
    
    \fill[blue!50] (2.38,2.34) rectangle (3.18,2.76);
    \fill[blue!30] (2.38,1.49) rectangle (3.18,1.91);
    \fill[blue!12] (2.38,0.64) rectangle (3.18,1.06);
    \node[font=\scriptsize] at (2.78,2.55) {CasTemp};
    \node[font=\scriptsize] at (2.78,1.70) {VNOIP};
    \node[font=\scriptsize] at (2.78,0.85) {CTCP};

    \draw[gray!35, rounded corners=2pt, fill=white] (3.6,0.25) rectangle (7.55,3.85);
    
    \node[anchor=west, font=\bfseries\scriptsize] at (3.75,3.65) {Shortcut Signatures};

    \fill[black] (3.78,3.25) circle (1.2pt);
    \node[anchor=west, font=\scriptsize] at (3.85,3.23) {Train--Test User Overlap on Taoke (\%)};
    \fill[blue!35] (4.85,2.85) rectangle (6.85,3.01); 
    \fill[blue!20] (4.85,2.65) rectangle (6.43,2.81);
    \fill[blue!5]  (4.85,2.45) rectangle (4.85,2.61);
    \node[anchor=east, font=\scriptsize] at (4.78,2.93) {Random};
    \node[anchor=east, font=\scriptsize] at (4.78,2.73) {Full};
    \node[anchor=east, font=\scriptsize] at (4.78,2.53) {Inductive};
    \node[anchor=west, font=\scriptsize] at (6.92,2.93) {79.9};
    \node[anchor=west, font=\scriptsize] at (6.50,2.73) {63.3};
    \node[anchor=west, font=\scriptsize] at (4.92,2.53) {0};

    \fill[black] (3.78,2.22) circle (1.2pt);
    \node[anchor=west, font=\scriptsize] at (3.85,2.20) {Performance Inflation on APS (\%)};
    \fill[orange!55] (4.85,1.82) rectangle (6.75,1.98);
    \fill[orange!35] (4.85,1.62) rectangle (6.52,1.78);
    \fill[orange!20] (4.85,1.42) rectangle (5.85,1.58);
    \node[anchor=east, font=\scriptsize] at (4.78,1.90) {CasFlow};
    \node[anchor=east, font=\scriptsize] at (4.78,1.70) {CasDo};
    \node[anchor=east, font=\scriptsize] at (4.78,1.50) {CasTemp};
    \node[anchor=west, font=\scriptsize] at (6.82,1.90) {151.3};
    \node[anchor=west, font=\scriptsize] at (6.59,1.70) {132.6};
    \node[anchor=west, font=\scriptsize] at (5.92,1.50) {80.5};

    \fill[black] (3.78,1.19) circle (1.2pt);
    \node[anchor=west, font=\scriptsize] at (3.85,1.17) {Temporal Drift Score};
    \fill[blue!45] (4.85,0.79) rectangle (6.15,0.95);
    \node[anchor=east, font=\scriptsize] at (4.78,0.87) {Weibo};
    \node[anchor=west, font=\scriptsize] at (6.22,0.87) {0.446};
    \fill[blue!55] (4.85,0.59) rectangle (6.27,0.75);
    \node[anchor=east, font=\scriptsize] at (4.78,0.67) {Twitter};
    \node[anchor=west, font=\scriptsize] at (6.34,0.67) {0.482};

\end{tikzpicture}
}
\caption{A compact preview of the benchmark story. \textbf{Left:} APS top-3 ranks change once the protocol is corrected from Random Cascade to Full Temporal / Inductive User. \textbf{Right-top:} Optimistic protocols preserve substantial train--test user overlap on Taoke. \textbf{Right-middle:} Performance on APS is significantly inflated under Random Cascade relative to Full Temporal. \textbf{Right-bottom:} Strong temporal drift on Twitter and Weibo demonstrates non-stationarity that undermines IID-based random splitting.}
\label{fig:intro_preview}
\end{figure}

This paper reorients the study of temporal cascade prediction around that question.
Instead of centering the narrative on a new state-of-the-art method, we build a benchmark-oriented framework for testing whether existing conclusions are robust under leakage-aware evaluation.
Concretely, the paper delivers a protocol suite, a conversion-rich benchmark asset, and a renewed evaluation standard for judging methods under forecasting-faithful conditions.

First, we formalize and correct the evaluation problem.
We begin with a toy study that exposes why random cascade partitioning can create temporally implausible supervision.
We then structure the study around three representative splits:
(i) \textbf{Random Cascade}, the conventional baseline that partitions cascades randomly without a global time boundary, allowing training and test sets to share future-era patterns and overlapping users;
(ii) \textbf{Full Temporal}, our advocated primary protocol that uses consecutive global time windows for observation and prediction, so that test supervision is drawn strictly from later periods than training supervision while still allowing long-lived items and users to persist across adjacent windows;
and (iii) \textbf{Inductive User}, the most stringent setting that combines the temporal cut with disjoint user sets across splits, eliminating identity-based shortcuts entirely.
This turns evaluation design into an object of study: methods should be judged not only by default-protocol accuracy, but also by how rankings, overlap exposure, and inflation change across the protocol suite.

Second, we broaden the benchmark assets with Taoke, a real-world e-commerce cascade dataset.
Taoke contains rich promoter and product features, complete forwarding traces, and downstream purchase conversions.
This supports both first-stage popularity forecasting and second-stage conversion forecasting in one benchmark.
We further use a larger internal extension, Taoke-large, that preserves the same semantics and task definitions while scaling up the product pool; its role is limited to operational stress testing rather than released benchmark comparison.

Third, we include CasTemp as a lightweight reference method.
CasTemp models self-cascade and cross-cascade temporal evidence through temporal walks, time-aware sequence encoding, and simple feature fusion.
Its role is intentionally modest: it is an efficient anchor method for interpreting the benchmark under stricter protocols and richer tasks.

Taken together, these pieces suggest that the central question is not simply which model wins on the legacy split, but which conclusions remain reliable once evaluation is made forecasting-faithful.
\Cref{fig:intro_preview} gives a compact preview aligned with our three-part framework.
\emph{(Left)} Protocol correction reshapes both ranks and apparent difficulty: on APS, the top method changes from CasFlow under Random Cascade to CasTemp under Full Temporal and Inductive User, showing that even the winner depends on the split definition.
\emph{(Right-top)} Leakage remains substantial under conventional protocols: on Taoke, Random Cascade and Full Temporal preserve 79.9\% and 63.3\% train--test user overlap respectively, whereas Inductive User removes it entirely.
\emph{(Right-middle)} This leakage translates into significant performance inflation: compared to Full Temporal, APS scores under Random Cascade are inflated by up to 151.3\% for CasFlow.
\emph{(Right-bottom)} The underlying data exhibits strong non-stationarity: the composite temporal-drift score reaches 0.482 on Twitter and 0.446 on Weibo, making random partitioning difficult to justify as a default.
Meanwhile, Taoke adds almost 3 million conversion events and rich side information, while the reference pipeline remains practical under the stricter benchmark, with 139$\times$--412$\times$ training speedups and 375$\times$--2457$\times$ inference speedups over the fastest baseline competitors on the released benchmark, and successful completion on a larger same-task internal extension where baseline systems run out of budget.

Overall, we first examine the evaluation protocols and show why benchmark conclusions can change once chronology and leakage channels are treated explicitly.
We then introduce the richer Taoke benchmark asset and its expanded task scope.
Finally, we use a lightweight reference pipeline to study what changes under the resulting benchmark specification, including cross-model ranking shifts, transfer behavior on the added conversion task, and the cost of maintaining a practical default standard.



Our contributions are summarized as follows.
\begin{itemize}
    \item \textbf{A protocol suite and renewed evaluation standard for temporal cascade prediction.}
    We identify the protocol sensitivity of current cascade benchmarks and reorganize evaluation around chronologically faithful splits, leakage diagnostics, inflation analysis, and temporal drift analysis.
    \item \textbf{A conversion-rich benchmark asset.}
    We introduce Taoke, a real-world cascade dataset with rich promoter/product features and downstream purchase conversions, enabling benchmark tasks beyond first-stage popularity prediction.
    \item \textbf{A lightweight reference pipeline.}
    We present CasTemp as a simple and efficient reference method that remains competitive under stricter evaluation while supporting scalable benchmarking.
\end{itemize}

\section{Related Work}\label{relatedwork}
\header{\textbf{Information Cascade Prediction.}}
Early work on popularity prediction relies on hand-crafted features—such as user profiles, structural depth, and temporal dynamics—combined with logistic regression or similar models \cite{szabo2010predicting,cheng2014can}. Later, RNN- and GRU-based methods model cascades as sequential events to capture temporal evolution \cite{cao2017deephawkes,liao2019popularity}, yet underutilized their inherent tree- or graph-structured topologies. This motivates graph-based approaches that represent cascades as evolving graphs and apply Graph Neural Networks (GNNs) for representation learning \cite{li2017deepcas,xu2021casflow}, effectively capturing local propagation patterns but typically treating cascades in isolation, ignoring inter-cascade interactions like competition. Recent models address this limitation: CTCP \cite{Lu2023ContinuousTimeGL} models cross-cascade correlations with memory updates, while CasDo \cite{cheng2024information} uses probabilistic diffusion models with ODEs \cite{hartman2002ordinary} to capture propagation uncertainty. However, these advances are achieved with the cost of increased architectural complexity and high computational overhead. 
Overall, early methods are structurally limited, while modern GNN-based models face efficiency and scalability challenges. 
Moreover, most follow a cascade-based data split \cite{xu2021casflow,Lu2023ContinuousTimeGL,cheng2024information,wang2026modeling} that does not temporally separate training and test data, creating a {time-travel} bias by leaking future information into training—leading to inflated and unrealistic performance estimates.

\header{\textbf{Dynamic Graph Learning.}}
Cascade propagation aligns naturally with continuous-time dynamic graphs—sequences of timestamped graph events \cite{yu2023towards,peng2025tidformer,peng2026gdgb}. Methods of dynamic graph learning model node representations for tasks like link prediction and node classification, including memory-based approaches \cite{kumar2019predicting,wang2021apan}, random walk-based methods \cite{starnini2012random,li2023zebra,lu2024improving}, and Transformer-based models \cite{yu2023towards,peng2025tidformer}. In popularity prediction, CTCP \cite{Lu2023ContinuousTimeGL} adopts memory mechanisms to update cascade and user states at each event. However, such methods suffer from computational bottlenecks due to frequent updates. 
Despite progress, efficiently adapting techniques from dynamic graph learning like temporal random walk and sequential modeling to cascade popularity prediction remains an open challenge.

\section{Preliminaries}\label{preliminary}

\header{\textbf{Information Cascade Graph.}}
The definition of conventional information cascade graphs naturally aligns with that of dynamic graphs. 
We consider a dynamic graph $G=(V,E,T)$ characterized by sets of nodes $V$, edges $E$, and timestamps $T$.
It captures the evolution of cascading diffusion through a sequence of chronologically ordered events $G=\{(src_i, tgt_i, c_i, t_i)\}_{i=1}^N$, with $0 \leq t_1 \leq \cdots \leq t_N$. 
Each event denotes the diffusion of a cascade $c_i$ (e.g., a post, paper, or product) from a source node $src_i \in V$ to a target node $tgt_i \in V$ at time $t_i \in T$. 
Nodes $src_i$ and $tgt_i$ are associated with features $\mathcal{N}_i^{src}, \mathcal{N}_i^{tgt} \in \mathbb{R}^{d_n}$, and the corresponding edge has a feature vector $\mathcal{E}_i^{t_i} \in \mathbb{R}^{d_e}$ encoding timestamped cascade features, where $d_n$ and $d_e$ denote the dimensions of node and edge embeddings, respectively.

\header{\textbf{Information Cascade Popularity Prediction.}}
We use $G^{c}(t)$ to denote the evolution process of cascade $c$ up to time $t$.
Given a cascade $c$ begins at $t_0^c$, after observing it for time $\Delta t_1$, the task is to predict its incremental cascade popularity $\Delta P = |G^{c}(t_0^c + \Delta t_2)|-|G^{c}(t_0^c + \Delta t_1)|$ over a future prediction window $\Delta t_2-\Delta t_1$.

\header{\textbf{Extended Cascade Graph with Popularity Conversions.}}
To support second-stage popularity conversion prediction, we extend the conventional information cascade graph by incorporating interactions between diffusion users and downstream users. 
Specifically, after each diffusion event from $src_j$ to $tgt_j$, the target diffusion user $tgt_j$ may trigger conversion events involving end users, such as likes, comments, or purchases, denoted as $(tgt_j, user_j, c_j, t_j)$. We collect these as a set $H = \{(tgt_j, user_j, c_j, t_j)\}_{j=1}^M$ with $0 \leq t_1 \leq \cdots \leq t_M$, forming an augmented view of the cascade dynamics that includes both propagation and conversion stages.

\header{\textbf{Second-stage Popularity Conversion Prediction.}}
We use $H^{c}(t)$ to denote the popularity conversion process of cascade $c$ until time $t$.
Given a cascade $c$ begins at $t_0^c$, after observing it for time $\Delta t_1$, the goal is to predict its incremental conversion count $\Delta C = |H^{c}(t_0^c + \Delta t_2)|-|H^{c}(t_0^c + \Delta t_1)|$ in the subsequent interval $\Delta t_2-\Delta t_1$.

\section{Benchmark Assets and Protocols}\label{dataset}

\subsection{Why the Existing Benchmark Needs Revision}\label{leakage}
\header{\textbf{Optimistic cascade-level splits.}}
Most prior cascade prediction papers adopt a random cascade split, typically assigning cascades to train/validation/test partitions without enforcing a global temporal boundary \cite{xu2021casflow,Lu2023ContinuousTimeGL,cheng2024information}.
This is problematic because the target itself is future growth.
When train and test cascades are sampled from overlapping periods, models may indirectly access synchronized bursts, overlapping users, or future-era activity patterns that would not be available in an actual forecasting pipeline.
For rigorous benchmarking, this matters because even a small amount of leakage can distort both absolute scores and relative model rankings.

\begin{table}[t]
\centering
\caption{Toy benchmark under the conventional cascade-based split. Lower is better.}
\vskip -0.1in
\label{tab:cas-based-toy}
\resizebox{0.4\textwidth}{!}
{
\begin{tabular}{l ccc ccc}
\toprule
\multirow{2}{*}{\textbf{Model}} & \multicolumn{3}{c}{\textbf{Scenario 1}} & \multicolumn{3}{c}{\textbf{Scenario 2}} \\
\cmidrule(lr){2-4} \cmidrule(lr){5-7}
 & \textbf{MSLE} & \textbf{MALE} & & \textbf{MSLE} & \textbf{MALE} & \\
\midrule
\rowcolor{lightpurple} MLP    & 6.3621 & 2.5498 & & 0.4510 & 0.6731 & \\
DeepHawkes & 5.5863 & 2.0954 & & 0.6449 & 0.7812 & \\
CasCN  & 5.4622 & 1.9477 & & 0.7831 & 0.8502 & \\
CasFlow& 5.8906 & 1.9985 & & 0.6251 & 0.9053 & \\
CTCP   & 5.1063 & 2.3519 & & 0.8546 & 0.9415 & \\
CasDo  & 6.0415 & 2.2980 & & 0.6831 & 0.9392 & \\
VNOIP  & 5.5802 & 2.0474 & & 0.6247 & 0.9010 & \\
\midrule
CasTemp  & 5.4821 & 2.0891 & & 0.7500 & 0.9044 & \\
\bottomrule
\end{tabular}
}
\end{table}

\begin{figure}[b]
    \centering
    \includegraphics[height=69mm]{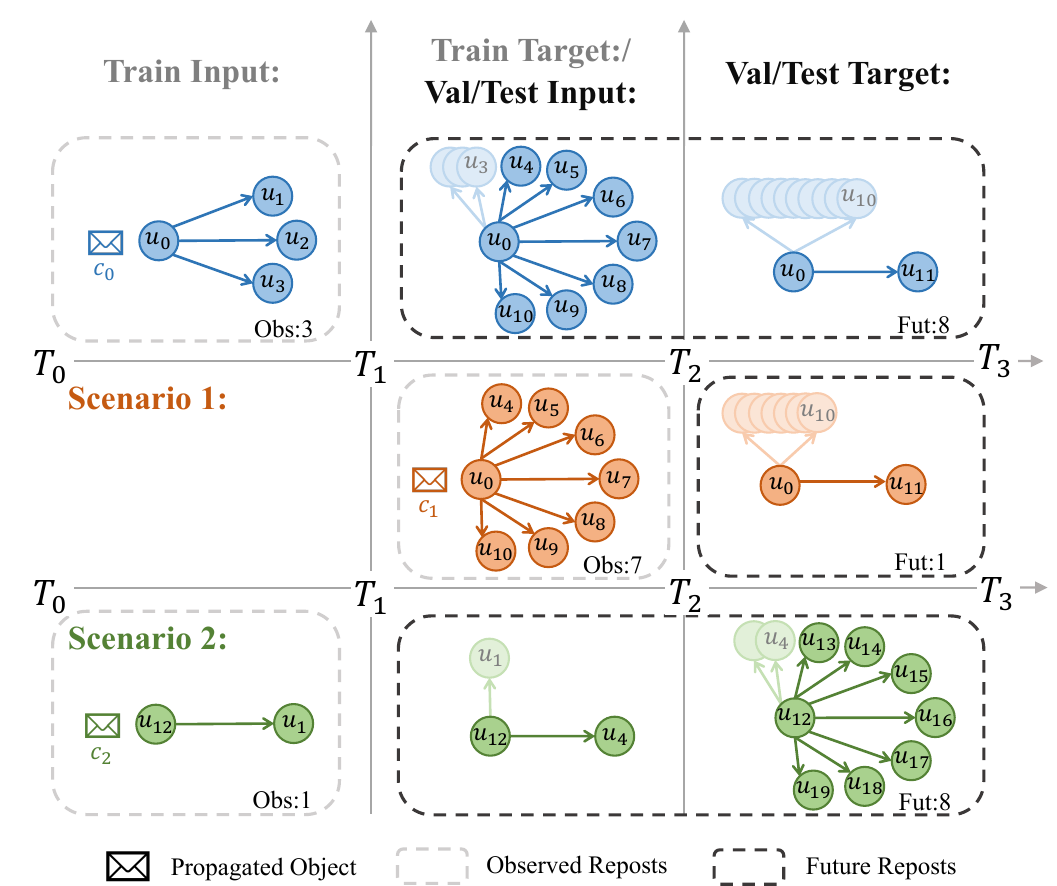}
    \vskip -0.1in
    \caption{Toy illustration of protocol sensitivity. Horizontally: conventional Random Cascade splitting with per-cascade observation and target windows. Vertically: a Full Temporal-style globally time-ordered split that separates train/validation/test by consecutive time segments.}
    \label{fig:toy_dataset}
\end{figure}

\header{\textbf{Toy evidence of protocol-induced shortcuts.}}
To make the issue concrete and intuitively verify the presence of information leakage under Random Cascade split, we construct a toy benchmark illustrated in \cref{fig:toy_dataset}.
It contrasts two evaluation views of the same diffusion process: the conventional Random Cascade-level partition and a Full Temporal-style globally time-ordered split.
Specifically, we set the observation window to one unit of time (gray dashed box) and define the incremental popularity up to the cutoff as the prediction target.
We design two scenarios to expose the failure mode:
In Scenario 1, the test cascade $c_1$ (orange) shares similar propagation dynamics with the training cascade $c_0$ (blue), including identical burst patterns between $T_1$--$T_2$ and decay phases from $T_2$--$T_3$.
Conversely, in Scenario 2, the test cascade $c_2$ (green) exhibits fundamentally different dynamics from $c_0$, originating from a distinct node ($u_{12}$ vs.\ $u_0$) with misaligned temporal phases.

Under genuine forecasting semantics, a model trained on $c_0$ should perform well on the similar $c_1$ but struggle with the dissimilar $c_2$.
However, under the Random Cascade protocol, future information is blurred into a shared pool, causing models to behave inconsistently with these intended semantics: they may exploit temporal shortcuts to perform poorly on apparently similar cases (Scenario 1) while achieving artificially high accuracy on dissimilar ones (Scenario 2).
Notably, even the simplest MLP achieves performance comparable to sophisticated baselines in \cref{tab:cas-based-toy}, suggesting that the apparent gains of complex models are largely driven by leaked signals rather than genuine predictive capacity.
We generate 1,000 synthetic cascades following this construction (70\%/30\% train-test split under the Random Cascade protocol).
The results in \cref{tab:cas-based-toy} should therefore be read not as a standalone benchmark, but as motivation for protocol revision, demonstrating that standard splits can invalidate evaluation even when test cascades appear semantically hard.

\begin{table}[t]
\centering
\caption{Toy benchmark under the corrected Full Temporal-style split. Lower is better.}
\vskip -0.1in
\label{tab:tem-order-toy}
\resizebox{0.4\textwidth}{!}
{
\begin{tabular}{l ccc ccc}
\toprule
\multirow{2}{*}{\textbf{Model}} & \multicolumn{3}{c}{\textbf{Scenario 1}} & \multicolumn{3}{c}{\textbf{Scenario 2}} \\
\cmidrule(lr){2-4} \cmidrule(lr){5-7}
 & \textbf{MSLE} & \textbf{MALE} & & \textbf{MSLE} & \textbf{MALE} & \\
\midrule
\rowcolor{lightpurple} MLP  & 2.9078 & 1.8136 & & 2.9884 & 1.5250 & \\
DeepHawkes & 1.6632 & 1.3905 & & 2.2501 & 1.3193 & \\
CasCN  & 2.9038 & 1.7192 & & 2.4778 & 1.5804 & \\
CasFlow& 4.6893 & 2.1320 & & 4.9673 & 2.0780 & \\
CTCP   & 1.9387 & 1.5913 & & 1.8842 & 1.3765 & \\
CasDo  & 5.6574 & 2.4400 & & 4.6387 & 2.1285 & \\
VNOIP  & 3.7326 & 1.9001 & & 2.0455 & 1.1901 & \\
\midrule
CasTemp  & 1.6104 & 1.4027 & & 1.8661 & 1.5412 & \\
\bottomrule
\end{tabular}
}
\end{table}

\begin{figure}[h]
    \centering
    \includegraphics[height=38mm]{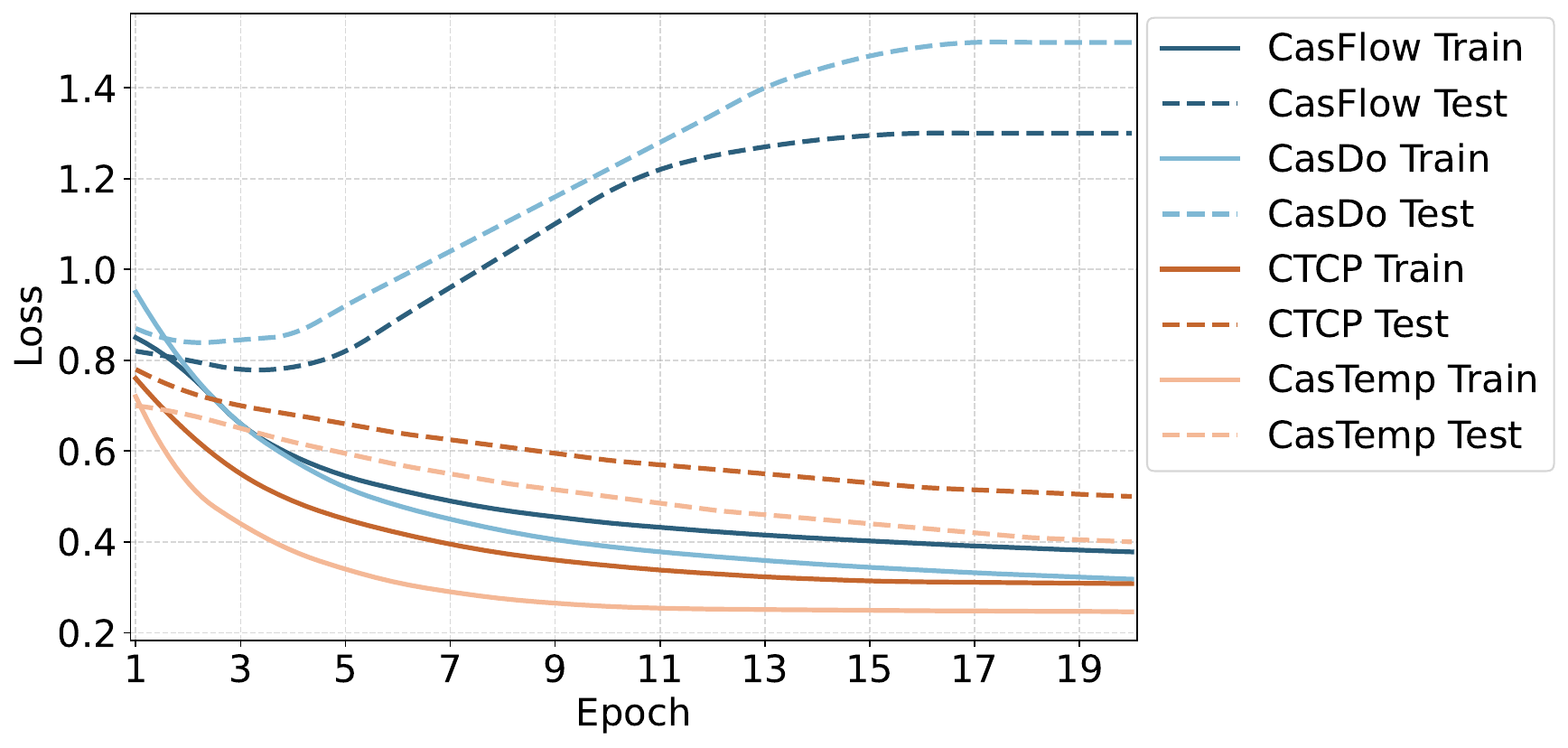}
    \vskip -0.1in
    \caption{Training and test loss under the time-ordered toy split.}
    \label{fig:train_test_loss}
\end{figure}

\header{\textbf{Correcting the split changes the qualitative conclusion.}}
The corrected toy results in \cref{tab:tem-order-toy} show the expected qualitative change.
Once the protocol is forecasting-consistent, over-parameterized models lose much of their apparent advantage, while lightweight methods generalize more stably; notably, several baselines now underperform even the simple MLP, whose relative standing improves markedly compared to \cref{tab:cas-based-toy}.
The loss curves in \cref{fig:train_test_loss} further suggest that some previously strong methods are sensitive to shortcut-driven overfitting.
This toy diagnosis is useful because it isolates the failure mode; the next subsection shows that the same issue reappears at benchmark scale.

\begin{figure}[b]
    \centering
    \includegraphics[height=31mm]{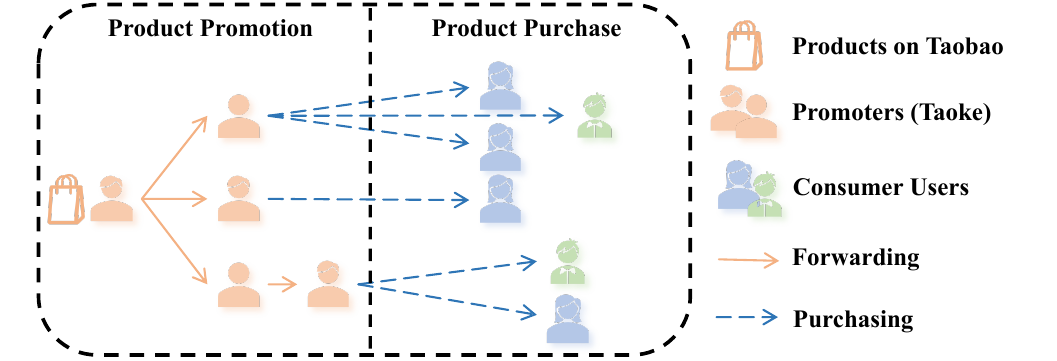}
    \caption{Illustration of the Taoke benchmark setting.}
    \label{fig:taoke_dataset}
\end{figure}

\begin{table}[b]
\centering
\caption{Statistical comparison across four benchmark datasets.}
\vskip -0.1in
\label{tab:stats}
\resizebox{0.48\textwidth}{!}
{
\begin{tabular}{lcccc}
\toprule
\textbf{Metric} & \textbf{Twitter} & \textbf{Weibo} & \textbf{APS} & \textbf{Taoke} \\
\midrule
Cascade   & 67,760 & 48,693 & 90,768 & 2,862 \\
Nodes   & 145,188 & 353,504 & 118,312 & 29,711 \\
Popularity Events  & 412,812 & 497,932 & 574,666 & 979,635 \\
Conversion Events  & \textbackslash & \textbackslash & \textbackslash & 2,990,747 \\
Avg. Path Length & 3.86 & 3.51 & 5.12 & 5.61 \\
Cascade Feature  & \ding{55} & \ding{55} & \ding{55} & \ding{51} \\
Node Feature  & \ding{55} & \ding{55} & \ding{55} & \ding{51} \\
\bottomrule
\end{tabular}
}
\end{table}

\subsection{Definitions of the Evaluation Protocols}\label{protocol-def}
\header{\textbf{Notation.}}
Let $\mathcal{C}$ denote the set of cascades and let $t_0^c$ be the creation time of cascade $c$.
For any time interval $I=[a,b)$, we write $G^c[I]$ for the cascade-$c$ subgraph induced by diffusion events whose timestamps lie in $I$, and
\[
N_c[I] = \left|\{(src_i,tgt_i,c_i,t_i): c_i=c,\; a \le t_i < b\}\right|
\]
for the number of cascade-$c$ diffusion events in that interval.
For protocols that split at the cascade level, every cascade uses the same local observation and target windows:
\[
x^c = G^c[t_0^c, t_0^c+\Delta t_1), \qquad
y^c = N_c[t_0^c+\Delta t_1, t_0^c+\Delta t_2).
\]
The distinction between protocols therefore lies in how the cascade set is partitioned and in which overlap channels remain accessible.

\header{\textbf{Protocol family.}}
We study three protocols.
\textbf{Random Cascade} partitions $\mathcal{C}$ uniformly at random into $\mathcal{C}_{\mathrm{tr}},\mathcal{C}_{\mathrm{va}},\mathcal{C}_{\mathrm{te}}$ and applies the local $(x^c,y^c)$ definition above.
For the recommended \textbf{Full Temporal} protocol, we choose global boundaries $\beta_0 < \beta_1 < \beta_2 < \beta_3 < \beta_4$ and define:
\[
\mathcal{D}^{\mathrm{FT}}_q =
\left\{
\bigl(G^c[\beta_{q-1},\beta_q),\; N_c[\beta_q,\beta_{q+1})\bigr)
:\;
c \in \mathcal{C},\;
N_c[\beta_{q-1},\beta_q) > 0
\right\},
\]
for $q \in \{1,2,3\}$, where $q=1,2,3$ correspond to train, validation, and test.
Finally, \textbf{Inductive User} keeps the same global windows but additionally requires:
\[
U(\mathcal{D}^{\mathrm{IU}}_{\mathrm{train}})
\cap
U(\mathcal{D}^{\mathrm{IU}}_{\mathrm{test}})
= \varnothing,
\]
where $U(\cdot)$ denotes the set of diffusion users appearing in a split.
In terms of benchmark role, Random Cascade serves as the optimistic historical baseline, Full Temporal is our main reporting protocol because it preserves forecasting semantics while remaining executable on all datasets, and Inductive User is best interpreted as a stress test for user-transfer reliance rather than as the default setting.

\subsection{Taoke: A Conversion-Rich Benchmark Dataset}\label{taoke}
\header{\textbf{Motivation.}}
Beyond protocol design, the benchmark also needs richer data.
Existing public cascade datasets usually contain propagation structure and timestamps but little contextual information, and they rarely support downstream conversion forecasting.
This limits both explanatory analysis and real-world relevance.

\begin{figure}[b]
    \centering
    \includegraphics[height=68mm]{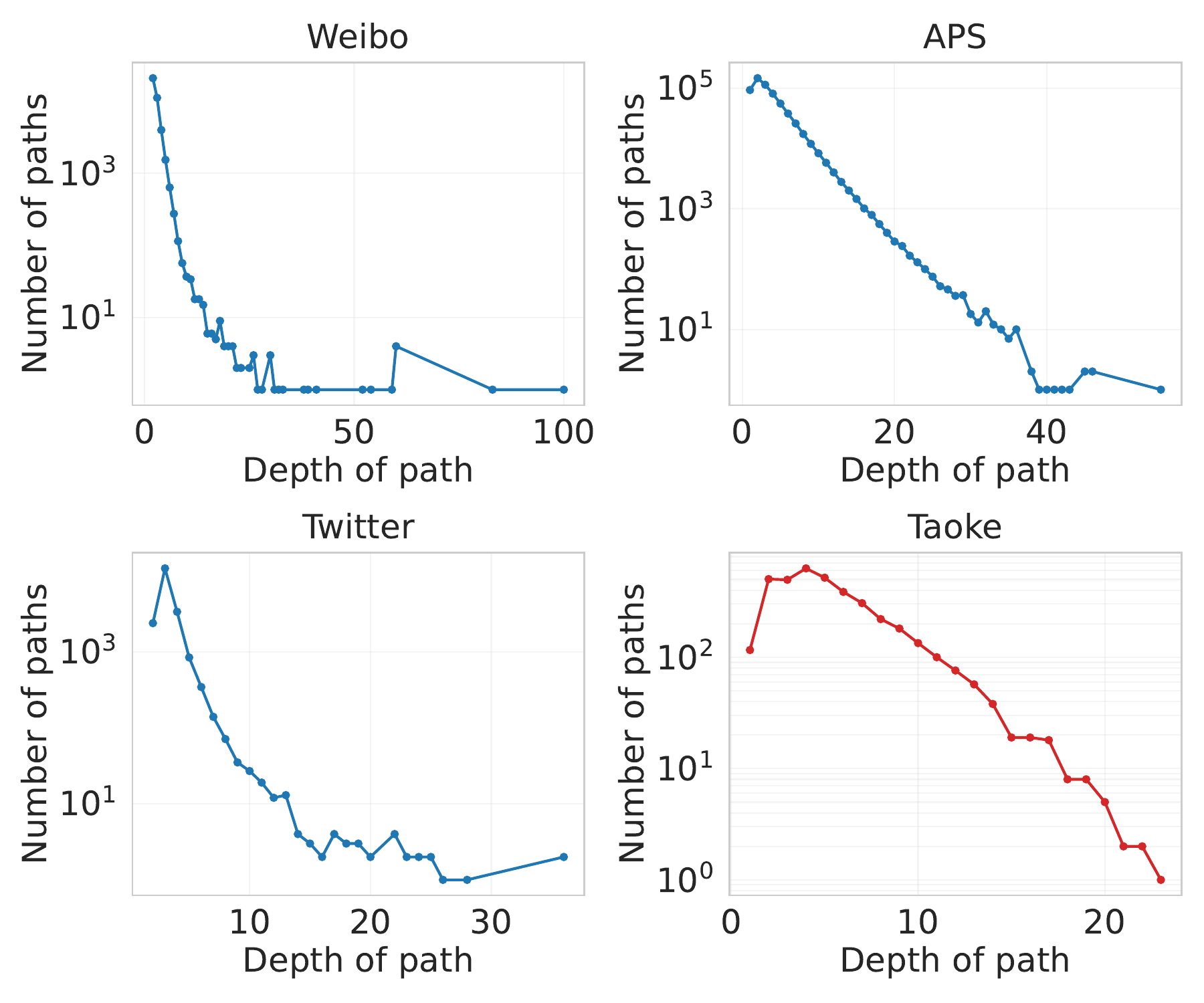}
    \vskip -0.1in
    \caption{Distribution of cascade-depth statistics across datasets.}
    \label{fig:depth}
\end{figure}

\header{\textbf{Dataset scope.}}
Taoke addresses this gap through a real e-commerce promotion scenario on Taobao.
As shown in \cref{fig:taoke_dataset}, each cascade corresponds to the diffusion of a promoted product among Taoke promoters, and the dataset additionally records conversion events from exposure to purchase.
This makes it possible to benchmark two related tasks within one asset: first-stage popularity forecasting and second-stage conversion forecasting.
Beyond the released benchmark asset, we also construct an internal large-scale extension, denoted \emph{Taoke-large}, that follows the same data semantics, feature schema, and task definitions as Taoke while covering a much larger product pool.
Because this extension is not publicly released and several baseline systems cannot complete end-to-end runs within the runtime budget, we use it only for efficiency stress testing rather than as a second full benchmark leaderboard.

Taoke also includes richer side information than the public baselines, including promoter features and product attributes such as price and commission rate.
These signals let us study not only how cascades spread, but also how business context interacts with propagation.
As shown in \cref{tab:stats}, Taoke complements the public datasets by retaining a comparable cascade structure while adding conversion labels and richer features.

The depth distribution in \cref{fig:depth} suggests that Taoke remains structurally consistent with known cascade behavior while introducing a more application-grounded setting.
This makes Taoke a benchmark asset that expands the experimental questions the field can ask.
Taoke-large should be interpreted as complementary evidence about scale.

\begin{figure*}[t]
\centering
\caption{Top-3 ranking matrix across three protocols. Darker shading indicates a better rank. `R', `F', and `I' denote Random Cascade, Full Temporal, and Inductive User, respectively. Full numerical results for all models are provided in \cref{tab:full_twitter,tab:full_aps,tab:full_weibo,tab:full_taoke}.}
\vskip 0.1in
\label{fig:rank_shift_heatmap}
\setlength{\tabcolsep}{4pt}
\renewcommand{\arraystretch}{1.05}
\begin{minipage}{0.24\textwidth}
\centering
\small
\textbf{Twitter}\par
\begin{tabular}{lccc}
\toprule
\textbf{Model} & \textbf{R} & \textbf{F} & \textbf{I} \\
\midrule
CasTemp    & \cellcolor{blue!50}1 & \cellcolor{blue!50}1 & \cellcolor{blue!50}1 \\
CasCN      & \cellcolor{blue!20}2 & \cellcolor{blue!20}2 & \cellcolor{blue!20}2 \\
DeepHawkes & \cellcolor{blue!8}3  & -- & -- \\
VNOIP      & -- & \cellcolor{blue!8}3 & -- \\
CTCP       & -- & -- & \cellcolor{blue!8}3 \\
\bottomrule
\end{tabular}
\end{minipage}\hfill
\begin{minipage}{0.24\textwidth}
\centering
\small
\textbf{Weibo}\par
\begin{tabular}{lccc}
\toprule
\textbf{Model} & \textbf{R} & \textbf{F} & \textbf{I} \\
\midrule
CasTemp    & \cellcolor{blue!50}1 & \cellcolor{blue!50}1 & \cellcolor{blue!50}1 \\
CasFlow    & \cellcolor{blue!20}2 & \cellcolor{blue!20}2 & -- \\
CasCN      & \cellcolor{blue!8}3  & -- & -- \\
DeepHawkes & -- & \cellcolor{blue!8}3 & \cellcolor{blue!8}3 \\
CTCP       & -- & -- & \cellcolor{blue!20}2 \\
\bottomrule
\end{tabular}
\end{minipage}\hfill
\begin{minipage}{0.24\textwidth}
\centering
\small
\textbf{APS}\par
\begin{tabular}{lccc}
\toprule
\textbf{Model} & \textbf{R} & \textbf{F} & \textbf{I} \\
\midrule
CasTemp & \cellcolor{blue!20}2 & \cellcolor{blue!50}1 & \cellcolor{blue!50}1 \\
CasFlow & \cellcolor{blue!50}1 & -- & -- \\
CasCN   & \cellcolor{blue!8}3 & \cellcolor{blue!8}3 & -- \\
VNOIP   & -- & \cellcolor{blue!20}2 & \cellcolor{blue!20}2 \\
CTCP    & -- & -- & \cellcolor{blue!8}3 \\
\bottomrule
\end{tabular}
\end{minipage}\hfill
\begin{minipage}{0.24\textwidth}
\centering
\small
\textbf{Taoke}\par
\begin{tabular}{lccc}
\toprule
\textbf{Model} & \textbf{R} & \textbf{F} & \textbf{I} \\
\midrule
CasTemp    & \cellcolor{blue!50}1 & \cellcolor{blue!50}1 & \cellcolor{blue!50}1 \\
CasFlow    & \cellcolor{blue!20}2 & \cellcolor{blue!20}2 & \cellcolor{blue!8}3 \\
CasCN      & \cellcolor{blue!8}3  & -- & -- \\
DeepHawkes & -- & \cellcolor{blue!8}3 & -- \\
CTCP       & -- & -- & \cellcolor{blue!20}2 \\
\bottomrule
\end{tabular}
\end{minipage}
\end{figure*}

\subsection{Analysis of the Evaluation Protocols}\label{bench_analysis}
\header{\textbf{Protocol sensitivity at benchmark scale.}}
The next question is whether the real benchmark behaves differently under these protocols.
The results in \Cref{fig:rank_shift_heatmap} show that it does.
For example, on APS, the best random-split baseline is CasFlow, but once chronology is enforced the winner changes to CasTemp and the entire runner-up group is reordered.
Twitter, Weibo, and Taoke keep the same winner across the three non-degenerate protocols, but their second and third positions still change materially, indicating that ranking stability cannot be assumed from a single split.

\header{\textbf{Leakage is multi-channel rather than scalar.}}
Different protocols suppress different shortcut channels.
Random Cascade removes item reuse across splits but still keeps non-trivial user overlap.
Full Temporal enforces chronological forecasting, yet it intentionally preserves item continuity because the same item may remain active across adjacent global windows; in fact, the train--test item overlap ratio is 1.0 on all four datasets under this protocol.
Inductive User removes train--test user reuse by construction.
\Cref{tab:protocol_stats_summary} quantifies these differences.

\begin{table}[t]
\centering
\caption{Protocol properties extracted from the actual split statistics. $\rho_u$ denotes train--test user overlap.}
\label{tab:protocol_stats_summary}
\resizebox{0.48\textwidth}{!}
{
\begin{tabular}{l c c c}
\toprule
\multirow{2}{*}{\textbf{Dataset}} & \multicolumn{3}{c}{\textbf{Train--Test User Overlap ($\rho_u$)}} \\
\cmidrule(lr){2-4}
& \textbf{Random Cascade} & \textbf{Full Temporal} & \textbf{Inductive User} \\
\midrule
Twitter & 0.257 & 0.285 & 0.000   \\
Weibo   & 0.170 & 0.209 & 0.000   \\
APS     & 0.072 & 0.175 & 0.000   \\
Taoke   & 0.799 & 0.633 & 0.000   \\
\bottomrule
\end{tabular}
}
\end{table}

\begin{table}[t]
\centering
\caption{APS case study for performance inflation when moving from Random Cascade to Full Temporal. Positive inflation means the random split reports an easier result.}
\label{tab:inflation_aps}
\resizebox{0.48\textwidth}{!}
{
\begin{tabular}{l c c c}
\toprule
\textbf{Model} & \textbf{Random Cascade} & \textbf{Full Temporal} & \textbf{Inflation} \\
\midrule
CasCN        & 1.192 & 2.496 & +109.3\% \\
CasDo        & 2.081 & 4.839 & +132.6\% \\
CasFlow      & 1.058 & 2.659 & +151.3\% \\
CasTemp      & 1.066 & 1.924 & +80.5\% \\
CTCP         & 1.356 & 2.802 & +106.7\% \\
DeepHawkes   & 1.431 & 2.524 & +76.4\% \\
VNOIP        & 1.740 & 2.399 & +37.9\% \\
\bottomrule
\end{tabular}
}
\end{table}

The preferred protocol need not minimize every scalar leakage statistic.
Because Full Temporal preserves forecasting-faithful item continuity, its aggregate leakage score is slightly higher than Random Cascade on Twitter, Weibo, and APS.
This reflects the benchmark design choice that removing temporal invalidity is more important than forcing all overlap channels to zero.
Moderate user overlap also reflects real deployment conditions where long-active users naturally span training and test periods.
Inductive User remains useful precisely because it isolates the user-transfer channel that Full Temporal intentionally leaves available.

\header{\textbf{Performance inflation exists, but its magnitude is dataset-dependent.}}
As shown in \cref{tab:inflation_aps}, inflation is quite striking on APS.
Relative to Full Temporal, Random Cascade improves every APS method, often by a large margin.
\Cref{tab:inflation_aps} is therefore our clearest case study of how optimistic protocols can materially overstate progress.

\begin{figure}[t]
\centering
\includegraphics[width=0.45\textwidth]{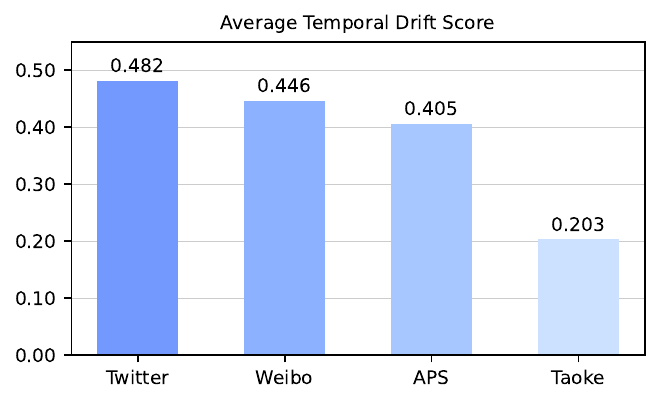}
\caption{Average adjacent-window temporal drift across datasets. Drift is computed as the mean of $(1 - \text{user Jacc.})$, $(1 - \text{cascade Jacc.})$, and burstiness JSD. Higher bars indicate stronger non-stationarity.}
\label{fig:drift_summary}
\end{figure}

\header{\textbf{Temporal drift further weakens the IID assumption.}}
Even if one ignores direct overlap contamination, the data are not temporally stationary.
\Cref{fig:drift_summary} summarizes adjacent-window drift across all four datasets.
Twitter and Weibo exhibit the highest average composite drift, while Taoke is more stable in aggregate but still far from IID: its user and cascade identities are persistent, yet burstiness and conversion-relevant activity still shift over time.
This is precisely the setting in which random cascade mixing of windows becomes hard to justify.

\begin{table}[t]
\centering
\caption{Drift profiles averaged over adjacent windows. User/cascade Jacc.\ measure identity continuity between consecutive windows, and burstiness JSD measures temporal-pattern shift in per-cascade burstiness distributions. Higher Jaccard indicates greater continuity; higher JSD indicates stronger pattern change.}
\label{tab:drift_profile}
\resizebox{0.48\textwidth}{!}
{
\begin{tabular}{c c c c c}
\toprule
\multirow{2}{*}{\textbf{Dataset}} 
& \multicolumn{3}{c}{\textbf{Per-Metric Drift Components}} \\
\cmidrule(lr){2-4}
& \textbf{User Jacc.} & \textbf{Cascade Jacc.} & \textbf{Burst. JSD} & \textbf{Avg. Drift} \\
\midrule
Twitter & 0.157 & 0.403 & 0.0049 & 0.482 \\
Weibo   & 0.101 & 0.564 & 0.0021 & 0.446 \\
APS     & 0.257 & 0.529 & 0.0005 & 0.405 \\
Taoke   & 0.498 & 0.916 & 0.0223 & 0.203 \\
\bottomrule
\end{tabular}
}
\end{table}

As shown in \cref{tab:drift_profile}, Twitter and Weibo have the weakest user continuity, with average adjacent-window user Jaccard scores of only 0.157 and 0.101, respectively, which helps explain why random mixing can easily blur distinct activity eras.
APS shows lower composite drift overall, but its user and cascade identities still turn over enough to violate a naive IID interpretation.
Taoke is different again: its user and cascade identities are substantially more persistent, yet its average burstiness JSD is the largest among the four datasets.
Looking back at the underlying window statistics, this comes from genuine popular business-cycle variation: the average cascade size rises from 51.8 in window 1 to 71.8 in window 4, while active users increase from 16.1K to 20.0K.
In other words, Taoke is temporally smoother in who participates, but still non-stationary in how strongly campaigns propagate.

\header{\textbf{What a revised benchmark must guarantee.}}
The evidence above suggests three design requirements.
First, the main protocol must be \emph{temporally faithful}: train, validation, and test should correspond to genuinely different forecasting eras rather than randomly mixed local windows.
Second, the benchmark should make \emph{leakage channels explicit}: user reuse, item reuse, and temporal co-movement should be measurable instead of hidden behind a single score.
Third, the benchmark must remain \emph{operationally executable}: the recommended setting should work across all released datasets instead of collapsing on validation or test partitions.
These requirements are what convert the discussion from an isolated critique of old splits into a benchmark specification.

\header{\textbf{Benchmark specification delivered by this paper.}}
The benchmark contribution of this paper has three layers: a \emph{benchmark asset} built from three standard public datasets together with Taoke, a \emph{protocol suite} built from Random Cascade, Full Temporal, and Inductive User, and a \emph{renewed evaluation standard} under which Full Temporal is the default reporting protocol and the others serve as diagnostics.
In short, the paper does not merely recommend a better split; it specifies what evidence should accompany future claims on temporal cascade prediction benchmarks.

\header{\textbf{Why Full Temporal is the main protocol.}}
Putting these observations together leads to a practical benchmark decision.
Random Cascade is too optimistic for trustworthy reporting.
Inductive User is valuable, but it changes the task by removing user transfer and is therefore better interpreted as a stress test than as the default setting.
We consequently adopt Full Temporal as the main protocol: it keeps the forecasting timeline faithful, remains executable on all datasets, and yields a single reporting setting whose conclusions do not depend on a degenerate split.
This choice is therefore not a convenience decision but part of the benchmark specification itself: Full Temporal defines the default comparison surface, while the other protocols clarify how fragile a conclusion would be if that default were ignored.

\section{Reference Method}\label{method}

We propose CasTemp, a {lightweight reference method}, for the benchmark.
It provides a simple but effective temporal pipeline that can be rerun across protocols, supporting both first-stage popularity and second-stage conversion prediction on Taoke, and exposing the preprocessing and update interfaces needed by the efficiency and maintenance studies.

\begin{figure*}[t]
    \centering
    \includegraphics[height=45mm]{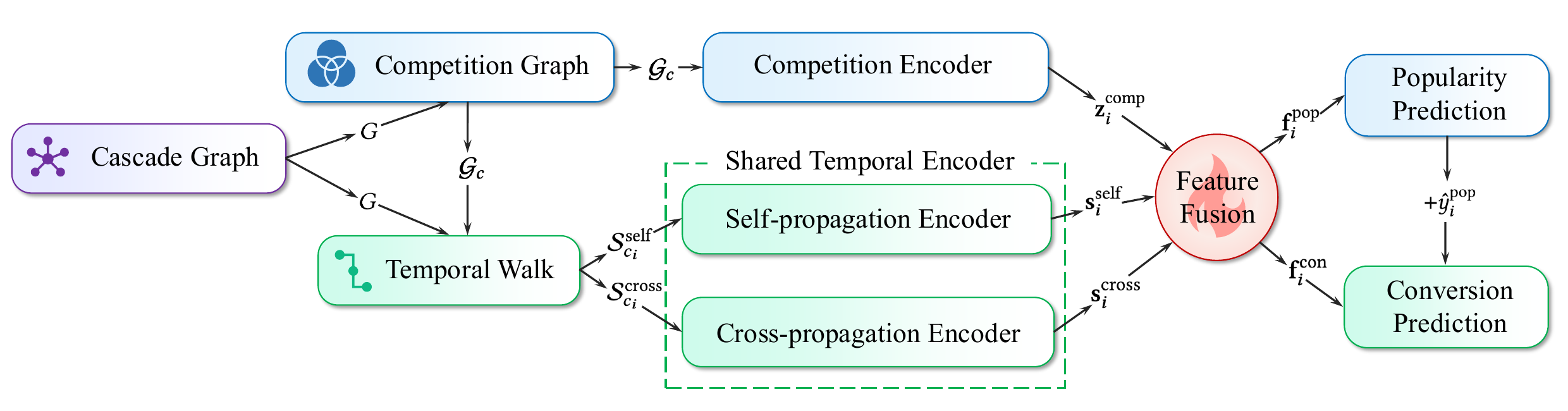}
    \vskip -0.1in
    \caption{CasTemp as a benchmark reference pipeline. It combines a cascade competition graph, self- and cross-cascade temporal sequences, and lightweight predictors for popularity and conversion forecasting.}
    \label{fig:pipeline}
\end{figure*}

\header{\textbf{Design principle.}}
Once the evaluation is made leakage-aware, the main benchmark need is no longer another highly specialized architecture.
What is more useful is a reference model that is easy to rerun, cheap to maintain, and still expressive enough to avoid becoming a straw-man baseline.
CasTemp follows that goal by combining a small amount of cross-cascade structure with recency-aware sequence encoding and shallow prediction heads.

\header{\textbf{Competition graph and temporal sequences.}}
Given the dynamic cascade graph in \cref{preliminary}, CasTemp first constructs an inter-cascade competition graph $\mathcal{G}_c = (\mathcal{C}, \mathcal{E}_c, \mathbf{w}_c)$.
Each node is a cascade, and the edge weight between cascades $c_i$ and $c_j$ is the Jaccard overlap of their promoter sets:
\[
w_{ij} = \frac{|U_i \cap U_j|}{|U_i \cup U_j|}.
\]
Only edges above a threshold $\tau_1$ are retained.
This graph provides a lightweight proxy for competition or collaboration among cascades and is simple to update through set maintenance.

For each target cascade $c_i$, CasTemp then extracts two ordered sequences: a self-propagation sequence $\mathcal{S}_{c_i}^{\text{self}}$ containing the recent events of $c_i$, and a cross-propagation sequence $\mathcal{S}_{c_i}^{\text{cross}}$ sampled from neighboring cascades in $\mathcal{G}_c$ by temporal random walks.
The former describes the cascade's own growth trajectory, while the latter captures nearby external activity without introducing a heavy graph-diffusion module.

\header{\textbf{Temporal sequence encoder.}}
Each event is represented by promoter features together with a sinusoidal encoding of normalized timestamps.
To reflect the stronger predictive value of recent behavior, CasTemp applies an exponential decay prior
\[
\gamma_k = \exp\left(-\lambda (t_{\max} - t_k)\right),
\]
where $t_{\max}$ is the latest timestamp in the sequence.
The decay-weighted events are encoded by a GRU \cite{dey2017gate} with additive attention, yielding $\mathbf{s}_i^{\mathrm{self}}$ and $\mathbf{s}_i^{\mathrm{cross}}$ for the self and cross sequences, respectively.
This keeps the temporal module compact while making the recency bias explicit.

\header{\textbf{Competition-aware cascade representation.}}
CasTemp further applies a lightweight GAT \cite{velivckovic2018graph} layer on $\mathcal{G}_c$ to produce a competition-aware cascade embedding $\mathbf{z}_i^{\mathrm{comp}}$.
This step is intentionally shallow: it is not meant to be a heavy graph learner, only a compact summary of the local inter-cascade context associated with $c_i$.

\header{\textbf{Popularity predictor.}}
For first-stage popularity forecasting, CasTemp concatenates the competition-aware embedding, the self- and cross-sequence summaries, historical popularity features, and dataset-specific side information:
\begin{equation}
    \mathbf{f}_i^{\mathrm{pop}} =
    \left[
    \mathbf{z}_i^{\mathrm{comp}};
    \mathbf{s}_i^{\mathrm{self}};
    \mathbf{s}_i^{\mathrm{cross}};
    \mathbf{h}_i^{\mathrm{his-pop}};
    \mathbf{h}_i^{\mathrm{fused}}
    \right].
\end{equation}
An MLP with Softplus output then predicts non-negative incremental popularity:
\[
\hat{y}_i^{\mathrm{pop}} = \mathrm{MLP}_{\mathrm{pop}}(\mathbf{f}_i^{\mathrm{pop}}).
\]
The simple head guarantees that the goal of testing whether a compact representation already suffices under the benchmark's stricter evaluation setting.

\header{\textbf{Conversion predictor.}}
On Taoke, the same reference pipeline is extended to second-stage conversion forecasting.
The conversion predictor reuses the same structural and temporal representations, then augments them with historical conversion signals and the predicted popularity:
\begin{equation}\label{eq:con_input}
    \mathbf{f}_i^{\mathrm{con}} =
    \left[
    \mathbf{z}_i^{\mathrm{comp}};
    \mathbf{s}_i^{\mathrm{self}};
    \mathbf{s}_i^{\mathrm{cross}};
    \mathbf{h}_i^{\mathrm{his-con}};
    \mathbf{h}_i^{\mathrm{fused}};
    \hat{y}_i^{\mathrm{pop}}
    \right].
\end{equation}
The final prediction is
\[
\hat{y}_i^{\mathrm{con}} = \mathrm{MLP}_{\mathrm{con}}(\mathbf{f}_i^{\mathrm{con}}).
\]
This keeps the second-stage task tightly coupled to the first-stage setup and makes Taoke useful as a benchmark for propagation-to-conversion modeling rather than an unrelated auxiliary task.

\header{\textbf{Why CasTemp fits this paper.}}
CasTemp is intentionally modest in scope.
It is simple enough that benchmark conclusions are not confounded by excessive architectural complexity, yet strong enough to remain competitive under Full Temporal evaluation and to support the added conversion task.
For a benchmarking study, that balance is more useful than over-centering a highly specialized method whose gains may be harder to disentangle from benchmark artifacts.

\section{Benchmark Evaluation and Reference Studies}\label{experiment}
The analyses in \cref{leakage,bench_analysis} establish why existing protocols are insufficient; they belong naturally to the benchmark-diagnosis narrative that precedes this experimental section.
What remains here are the evaluations that presuppose a sound protocol: the Full Temporal split, the richer Taoke task, and the systems-oriented behavior of a lightweight reference pipeline.
The emphasis is on cross-model insight under the renewed benchmark standard: which ranking patterns persist, which disappear, and what additional properties become visible once evaluation is forecasting-faithful.

\subsection{Experimental Setup}
\header{\textbf{Datasets.}}
We evaluate on four real-world datasets: Twitter \cite{weng2013viralitytwitter}, Weibo \cite{cao2017deephawkes}, APS \footnote{\url{https://journals.aps.org/datasets}}, and Taoke.
\cref{tab:stats} summarizes their statistics, while \cref{app:dataset} gives additional details.
For Full Temporal split, the four equal-duration windows correspond to 2 days on Twitter, 1 hour on Weibo, 5 years on APS, and 1 day on Taoke.

\header{\textbf{Reporting policy.}}
To keep the experimental section focused, we use the protocol family asymmetrically.
Random Cascade and Inductive User are used earlier in \cref{dataset} to diagnose benchmark sensitivity and leakage channels.
The default comparisons in this section are therefore reported under Full Temporal, because it is forecasting-faithful, non-degenerate across all four datasets, and directly comparable across model families.

\header{\textbf{Models.}}
Our main benchmark covers representative methods from several model families: DeepHawkes \cite{cao2017deephawkes}, CasCN \cite{chen2019information}, CasFlow \cite{xu2021casflow}, CTCP \cite{Lu2023ContinuousTimeGL}, CasDo \cite{cheng2024information}, VNOIP\cite{wang2026modeling}, and CasTemp.
The toy analysis in \cref{leakage} additionally uses an MLP sanity-check baseline.
This breadth is important because the main question is not whether one method wins on a single split, but whether conclusions are stable across protocols, especially for a benchmarking paper.

\header{\textbf{Metrics.}}
For first-stage popularity prediction, we use Mean Squared Logarithmic Error (MSLE) and Mean Absolute Logarithmic Error (MALE). For second-stage conversion prediction, we additionally adopt Hit@40—a commonly used metric in information retrieval—which measures the proportion of predictions whose error relative to the ground truth is less than 40\%. 

\header{\textbf{Implementation.}}
All models are trained for up to 100 epochs with early stopping patience of 30, and any single run exceeding 48 hours is forcibly terminated and evaluated at its last checkpoint.
Baselines are integrated from their original open-source implementations with published hyperparameters into our unified codebase to ensure fair comparison.
CasTemp is trained with learning rate 0.01; the sequence encoder uses hidden dimension 16, temporal decay coefficient $\lambda=0.1$, self-sequence length $L_{\max}=10$, competition-graph threshold $\tau_1=0.1$, and temporal random walk budget $(\tau_2,\tau_3)=(5,10)$.
These settings are chosen to balance efficiency and stability and are further diagnosed in \cref{exp:reference-diagnostics}.
All methods are evaluated on the same released splits and reporting windows.
Experiments are conducted on an NVIDIA RTX A6000 48GB GPU.

\subsection{Performance Under the Main Protocol}\label{exp:reference-performance}
\Cref{tab:pop_results_full} reports popularity-forecasting results on the four datasets under Full Temporal.
The interpretation is intentionally benchmark-centric: Full Temporal does not support a universal winner among existing model families, but instead exposes a heterogeneous picture in which strengths depend strongly on dataset conditions.

\begin{table}[t]
\centering
\caption{Popularity prediction performance (MSLE, \textdownarrow) on four datasets under the Full Temporal protocol. Best results are bolded.}
\vskip -0.1in
\label{tab:pop_results_full}
\resizebox{0.49\textwidth}{!}
{
\begin{tabular}{l c c c c}
\toprule
\textbf{Method} & \textbf{Twitter} & \textbf{Weibo} & \textbf{APS} & \textbf{Taoke} \\
\midrule
DeepHawkes   & $1.270 \pm 0.037$ & $1.729 \pm 0.005$ & $2.524 \pm 0.053$ & $3.363 \pm 0.132$ \\
CasCN        & $1.194 \pm 0.020$ & $2.587 \pm 0.517$ & $2.496 \pm 0.029$ & $3.432 \pm 0.054$ \\
CasFlow      & $1.240 \pm 0.007$ & $1.633 \pm 0.022$ & $2.659 \pm 0.035$ & $2.288 \pm 0.040$ \\
CTCP         & $1.447 \pm 0.002$ & $1.883 \pm 0.002$ & $2.802 \pm 0.016$ & $8.384 \pm 0.021$ \\
CasDo        & $2.101 \pm 0.030$ & $2.143 \pm 0.461$ & $4.839 \pm 0.201$ & $27.166 \pm 0.172$ \\
VNOIP        & $1.221 \pm 0.053$ & $2.478 \pm 0.061$ & $2.399 \pm 0.083$ & $6.574 \pm 0.105$ \\
\midrule
\rowcolor{lightpurple} CasTemp & $\mathbf{1.168 \pm 0.003}$ & $\mathbf{1.478 \pm 0.009}$ & $\mathbf{1.924 \pm 0.016}$ & $\mathbf{0.798 \pm 0.040}$ \\
\bottomrule
\end{tabular}
}
\end{table}

\begin{table}[t]
\centering
\caption{Second-stage conversion prediction performance (\textdownarrow) on Taoke under the Full Temporal protocol. Best results are bolded. CasTemp-W/O P removes $\hat{y}_i^{\mathrm{pop}}$; CasTemp-W/ GTP uses oracle first-stage popularity for analysis only.}
\vskip -0.1in
\label{tab:con_results}
\resizebox{0.46\textwidth}{!}
{
\begin{tabular}{l c c c}
\toprule
\multirow{2}{*}{\textbf{Method}} &
\multicolumn{3}{c}{\textbf{Taoke (Conversion)}} \\
\cmidrule(lr){2-4}
& \textbf{MSLE} & \textbf{MALE} & \textbf{Hit@40} \\
\midrule
DeepHawkes   & $42.515 \pm 1.601$& $5.912 \pm 0.235$& $2.10\% \pm 0.07\%$\\
CasCN        & $8.380 \pm 0.119$& $2.158 \pm 0.068$& $18.42\% \pm 1.41\%$\\
CasFlow      & $18.992 \pm 0.489$& $3.071 \pm 0.097$& $12.06\% \pm 1.37\%$\\
CTCP         & $10.494 \pm 0.105$& $2.489 \pm 0.040$& $16.23\% \pm 0.99\%$\\
CasDo        & $35.588 \pm 1.418$& $5.033 \pm 0.219$& $2.11\% \pm 0.13\%$\\
VNOIP        & $9.339 \pm 0.593$& $2.289 \pm 0.095$& $16.83\% \pm 0.60\%$\\
\midrule
\rowcolor{lightpurple} CasTemp & $\mathbf{1.955 \pm 0.120}$ & $\mathbf{0.764 \pm 0.023}$ & $\mathbf{54.60\% \pm 0.85\%}$ \\
CasTemp-W/O P & $3.506 \pm 0.331$ & $1.237 \pm 0.100$ & $45.69\% \pm 1.22\%$ \\
CasTemp-W/ GTP & $1.863 \pm 0.104$ & $0.711 \pm 0.046$ & $53.68\% \pm 0.83\%$ \\
\bottomrule
\end{tabular}
}
\end{table}

Four observations are worth highlighting.
First, the stricter protocol compresses some of the apparent separation among public-dataset baselines.
On Twitter, for example, the gap between the best baseline CasCN and CasTemp is small, which is consistent with the earlier diagnosis that forecasting-faithful evaluation removes part of the room for shortcut-driven gains.
Second, the relative ordering of baseline families is highly dataset-dependent even after the protocol is fixed.
VNOIP is the strongest non-CasTemp model on APS, CasCN is strongest on Twitter, and CasFlow is strongest on Weibo and Taoke.
This reveals that different datasets reward different inductive biases.
Third, the larger margins on APS and especially Taoke indicate that chronological fidelity does not only reduce inflated gains; it can also expose where temporal context and cross-cascade evidence matter more substantively.
Relative to the best non-CasTemp baseline, the error drops from 2.399 to 1.924 on APS and from 2.288 to 0.798 on Taoke.
Fourth, the combination of rank dispersion and margin variation argues against treating any one dataset as a proxy for the entire problem class.
The meaningful question is therefore not ``who is best overall,'' but which model families remain robust across domains once optimistic evaluation channels are removed.
These observations motivate the next subsection, which examines the richer Taoke task directly.

\begin{table}[b]
\centering
\caption{Scale of the internal Taoke-large extension. It preserves the same semantics, features, and task definitions as Taoke, but is used only for large-scale stress testing.}
\vskip -0.1in
\label{tab:taoke_large_stats}
\resizebox{0.35\textwidth}{!}{
\begin{tabular}{lcc}
\toprule
\textbf{Metric} & \textbf{Taoke} & \textbf{Taoke-large} \\
\midrule
Cascades & 2,862 & 126,515 \\
Nodes & 29,711 & 416,880 \\
Popularity events & 979,635 & 2,496,732 \\
Conversion events & 2,990,747 & 16,699,943 \\
\bottomrule
\end{tabular}
}
\end{table}

\subsection{Taoke as a Benchmark for Second-Stage Conversion}
\label{exp-second}
Taoke is not only a new dataset but also a task extension for the benchmark.
\cref{tab:con_results} evaluates second-stage conversion forecasting under the same time-aware split.
All methods reuse the embeddings learned for first-stage popularity forecasting and only train a new predictor for conversions, which makes the comparison a test of representation transfer rather than a separate end-to-end redesign.

The conversion results reinforce the value of Taoke as a benchmark asset.
The setting exposes a practical target that classic cascade datasets cannot express, and it tests whether first-stage cascade representations remain useful for downstream monetization-oriented prediction.
CasTemp performs strongly here, but the more important benchmark takeaway is that richer tasks alter the ranking criteria themselves: a model is now judged not only by how well it predicts spread, but also by whether its representation transfers to economically meaningful outcomes.
Methods that remain competitive on first-stage popularity do not transfer equally well to conversion forecasting, and the resulting ranking is more sharply separated than on the popularity task.
Relative to the best non-CasTemp baseline, CasTemp reduces MSLE from 8.380 to 1.955 and raises Hit@40 from 18.42\% to 54.60\%.
That gap is large enough to matter qualitatively and illustrates the point of adding Taoke to the benchmark: the evaluation target changes, and so does what counts as a strong model.

The ablation rows in \cref{tab:con_results} clarify the role of $\hat{y}_i^{\mathrm{pop}}$ in \cref{eq:con_input}.
Removing this signal causes a large deterioration across all three metrics, indicating that first-stage popularity is a meaningful bridge to second-stage conversion.
Using oracle popularity changes the result only marginally, which suggests that most of the gain comes from exposing a calibrated first-stage signal at all, not from assuming access to unattainable future information.

\begin{table*}[t]
\centering
\caption{Efficiency summary under Full Temporal. For the released benchmark datasets, the baseline shown is the fastest non-CasTemp method under the corresponding metric, and speedup is computed relative to that baseline. Taoke-large is an internal non-public extension with the same task semantics as Taoke but larger scale; it is reported only as a stress test, and OOT denotes the training time over 48 hours.}
\vskip -0.1in
\label{tab:efficiency_summary}
\resizebox{0.75\textwidth}{!}{%
\renewcommand{\arraystretch}{1.2}
\begin{tabular}{l c c c c c c}
\toprule
\textbf{Dataset} & \textbf{Train} & \textbf{Baseline} & \textbf{Train Ref.} & \textbf{Infer} & \textbf{Baseline} & \textbf{Infer Ref.} \\
\midrule
Twitter & 44.58s & CTCP & 6219.47s / 139.5$\times$ & 0.0185s & CTCP & 19.76s / 1066$\times$ \\
Weibo   & 13.18s & CTCP & 5426.36s / 411.8$\times$ & 0.0089s & CTCP & 14.33s / 1615$\times$ \\
APS     & 29.04s & CTCP & 6908.98s / 237.9$\times$ & 0.0135s & CTCP & 33.09s / 2457$\times$ \\
Taoke   & 8.30s  & CasCN & 1390.60s / 167.6$\times$ & 0.0104s & CasCN & 3.91s / 374.9$\times$ \\
Taoke-large & 1504.30s & N/A & All baselines OOT & 30.5532s & N/A & All baselines OOT \\
\bottomrule
\end{tabular}%
}
\end{table*}

\subsection{Efficiency and Scalability}
\begin{figure}[b]
    \centering
    \includegraphics[height=38mm]{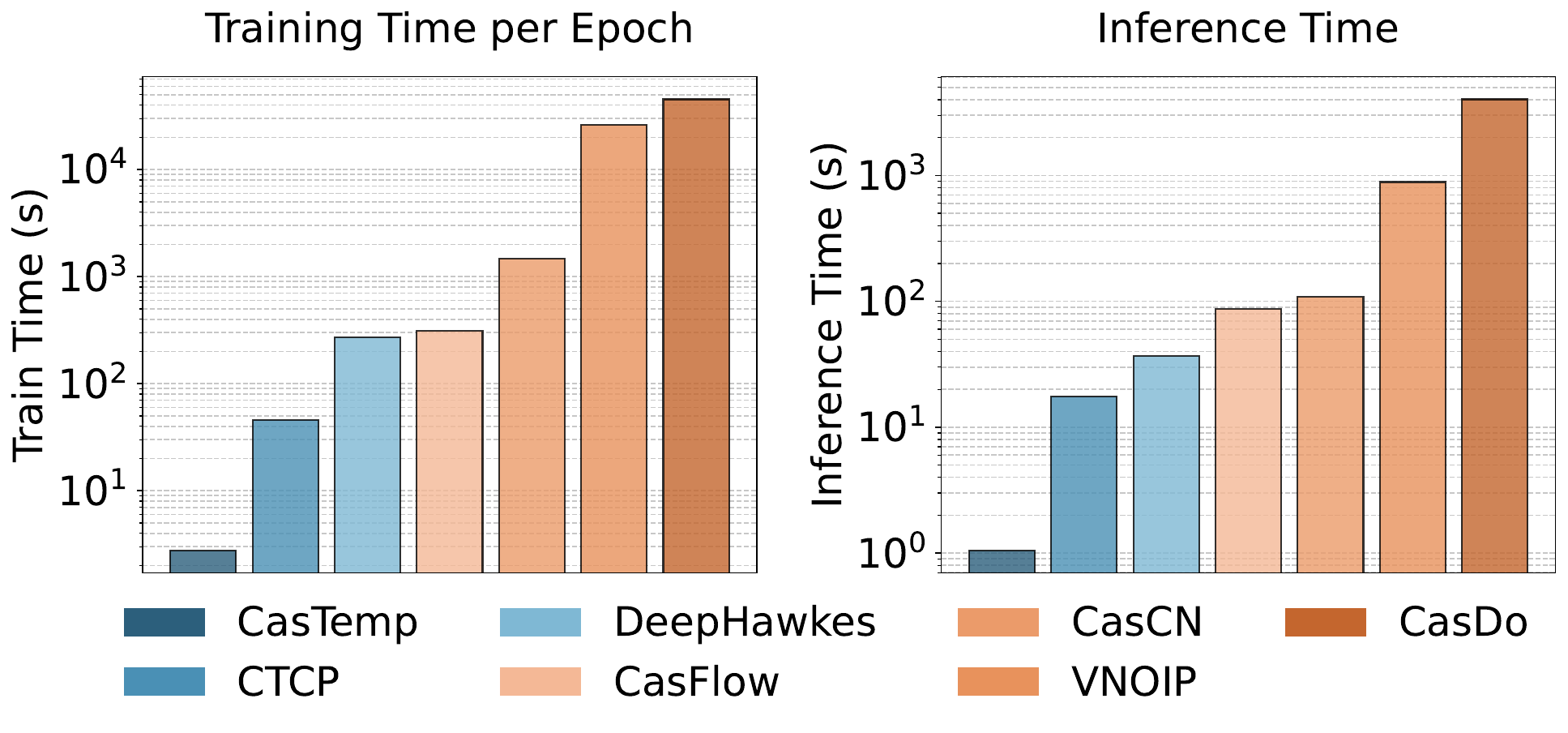}
    \caption{Training time per epoch and inference time of each method on Twitter.}
    \label{fig:train_infer_time}
\end{figure}

\Cref{tab:efficiency_summary} summarizes two kinds of systems evidence.
For the four released benchmark datasets, it reports the identity-run efficiency comparison between CasTemp and the fastest baseline under the same protocol.
It additionally includes Taoke-large, an internal large-scale extension with the same task semantics as Taoke but substantially larger scale, to assess whether the reference pipeline remains runnable when the product pool grows well beyond the released benchmark.
Its scale relative to the released Taoke benchmark is summarized in \cref{tab:taoke_large_stats}.
On the released datasets, the reference pipeline is between 140$\times$ and 412$\times$ faster to train, and between 375$\times$ and 2457$\times$ faster to infer, than the fastest baseline competitor under the same protocol.
Together with \cref{fig:train_infer_time}, this shows that the efficiency advantage is structural rather than incidental.
For a benchmark paper, this matters because a renewed evaluation standard must remain practical.
Taoke-large strengthens that claim from a stress-test perspective: CasTemp still finishes, whereas all baseline systems exceed the 48-hour training budget and are therefore marked OOT rather than compared as a full leaderboard.

\begin{table}[t]
\centering
\caption{CasTemp scalability on Taoke under synthetic event expansion.}
\vskip -0.1in
\label{tab:scalability_taoke}
\resizebox{0.48\textwidth}{!}
{
\begin{tabular}{c c c c c}
\toprule
\textbf{Event ratio} & \textbf{Events (M)} & \textbf{Preprocess (s)} & \textbf{Train (s)} & \textbf{Infer (s)} \\
\midrule
$2\times$  & 1.96 & 12.37 & 8.58 & 0.0109 \\
$4\times$  & 3.92 & 19.54 & 9.03 & 0.0111 \\
$8\times$  & 7.84 & 34.58 & 9.37 & 0.0118 \\
$16\times$ & 15.67 & 63.65 & 9.99 & 0.0129 \\
\bottomrule
\end{tabular}
}
\end{table}

The synthetic-expansion results in \cref{tab:scalability_taoke} complement the cross-model timing comparison.
They isolate the scaling behavior of CasTemp itself by expanding Taoke from 1.96M to 15.67M selected events.
Preprocessing grows from 12.37s to 63.65s, which is consistent with the expected increase in input volume, while train and inference remain in the narrow ranges.
This suggests that, for CasTemp, the dominant cost is not the predictor itself but the data preparation around it.
This separation is important: the identity-run table asks whether CasTemp is a practical benchmark anchor today, while the synthetic-expansion table asks whether the same pipeline remains usable as the event stream grows.

\subsection{Ablation Study}\label{exp:reference-diagnostics}
\begin{figure*}[t]
    \centering
    \includegraphics[height=34mm]{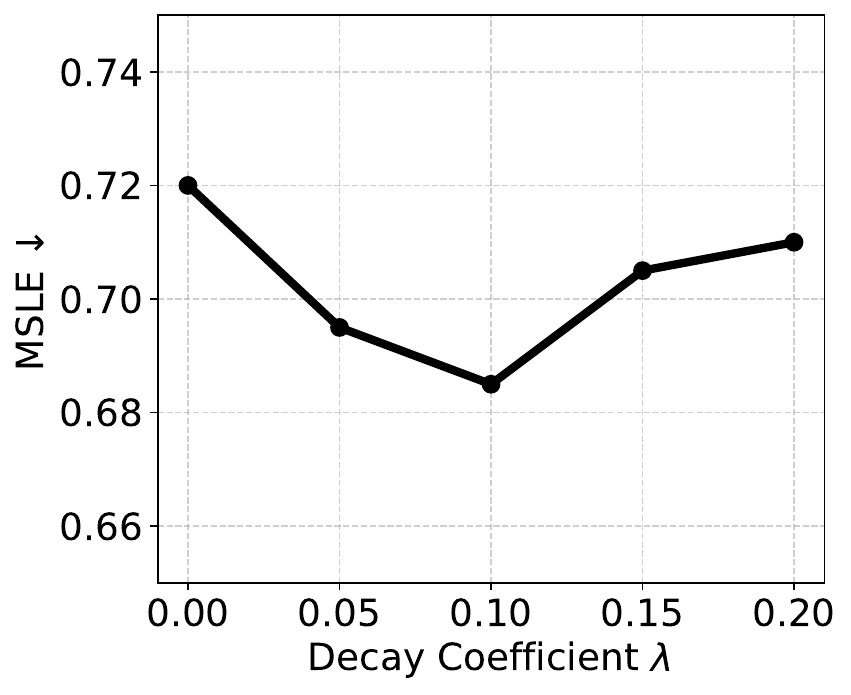}
    \includegraphics[height=34mm]{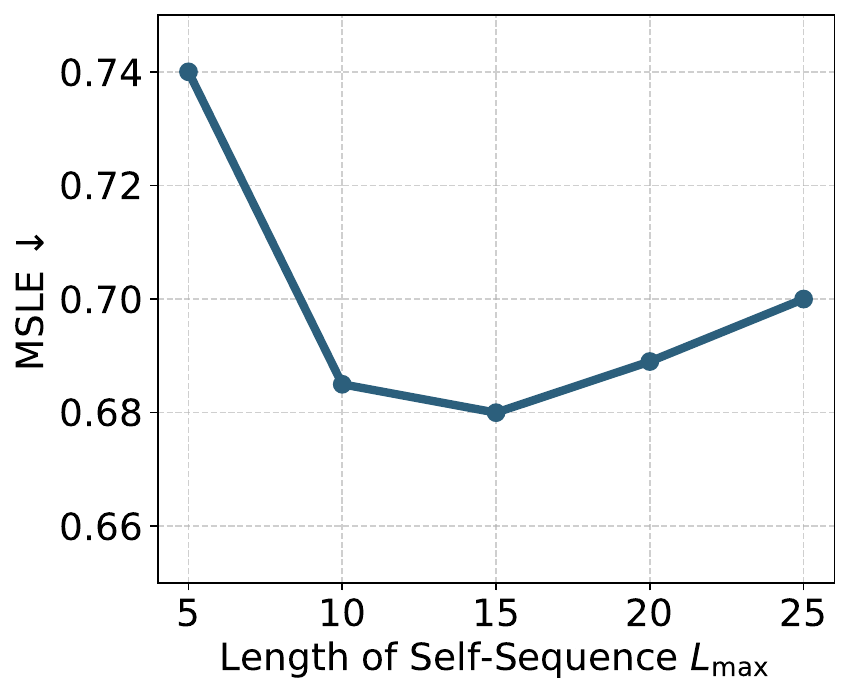}
    \includegraphics[height=34mm]{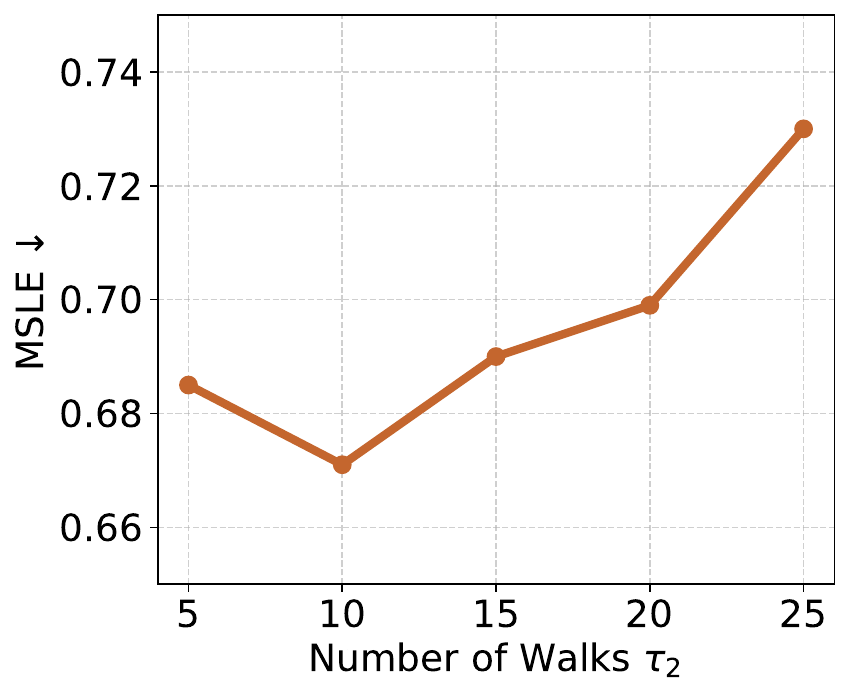}
    \includegraphics[height=34mm]{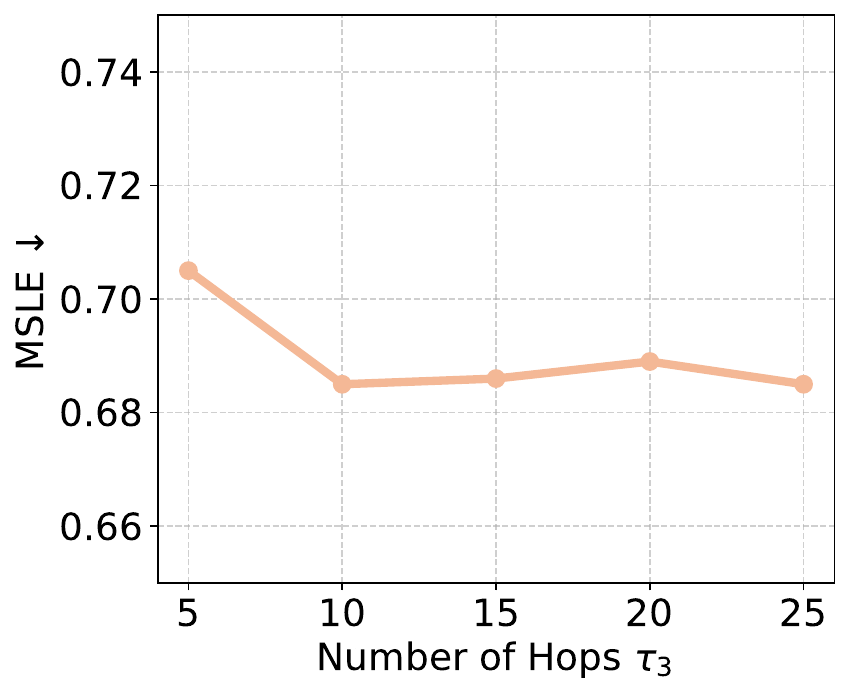}
    \caption{Hyperparameter analysis of the CasTemp reference method on Taoke.}
    \label{fig:hyper}
\end{figure*}

\cref{fig:hyper} studies the sensitivity of four core hyperparameters on CasTemp.
The results suggest that the method is reasonably stable across several nearby settings, with $\lambda=0.1$ and $(L_{\max},\tau_2,\tau_3)=(10,5,10)$ providing a good efficiency-accuracy trade-off.
This matters because it indicates that the reference method is not winning only through brittle hyperparameter tuning.

\begin{figure}[t]
    \centering
    \includegraphics[height=62mm]{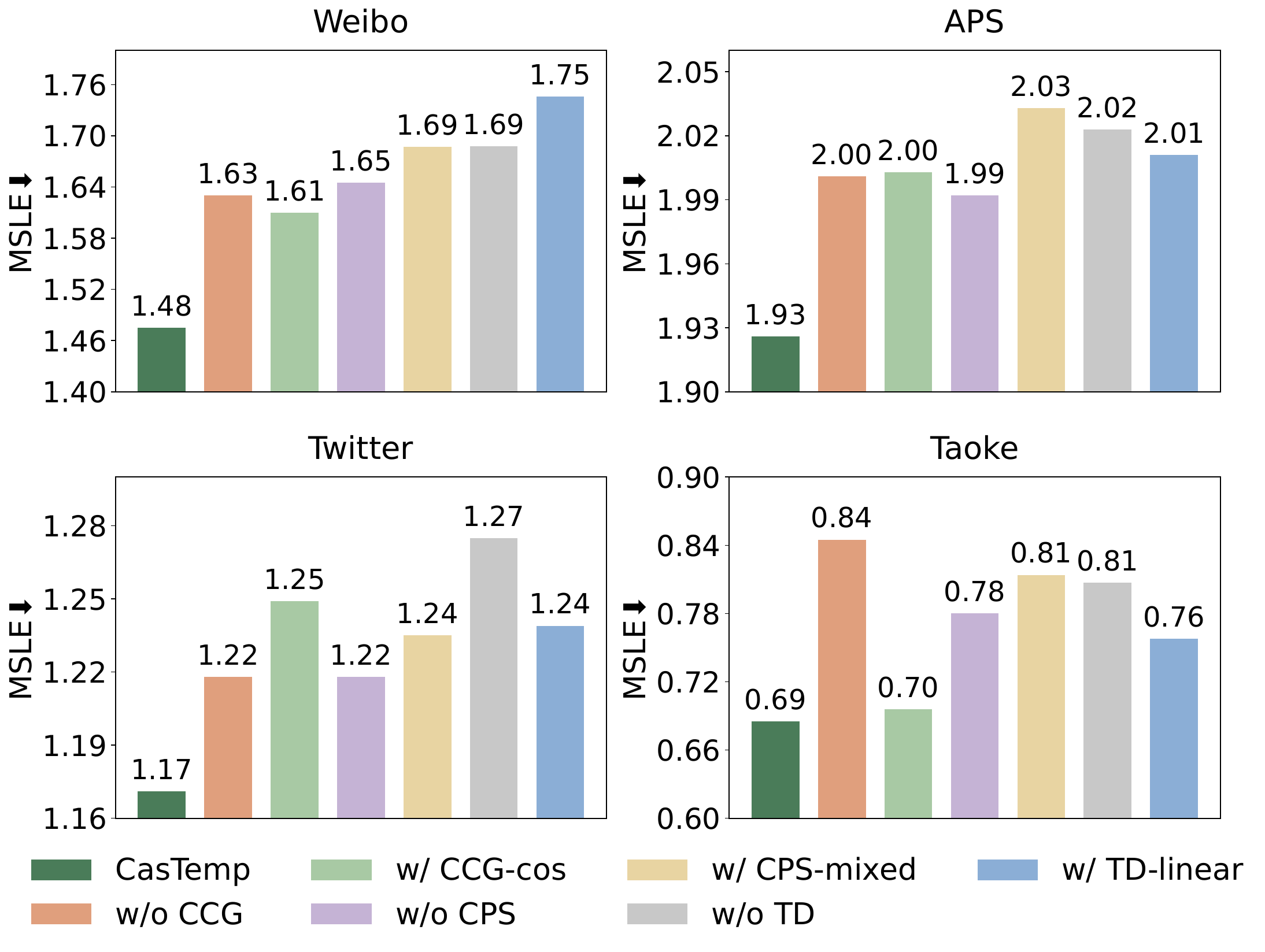}
    \caption{Ablation of key components in the CasTemp reference method.}
    \label{fig:ab}
\end{figure}

We also report a component-level ablation in \cref{fig:ab}.
Its role is supportive rather than central: it confirms that the competition graph, cross-cascade sequence construction, and temporal decay mechanism each contribute non-trivially to the reference pipeline's effectiveness, so the model's behavior is not driven by one isolated trick.

\subsection{Discussion}
Taken together, the experiments reposition temporal cascade prediction as a benchmark design problem as much as a modeling problem.
Several cross-model insights now emerge more clearly.
Reliable evaluation does not preserve a single, globally dominant baseline family; instead, rankings become more dataset-specific once leakage channels are controlled.
Expanding the asset from popularity forecasting to conversion forecasting further changes the comparison surface by rewarding transferability, not only one-shot predictive fit.
From a systems perspective, the benchmark also remains runnable at practical cost, which is necessary if the proposed protocol suite and reporting standard are to function as shared infrastructure rather than as one-off analysis artifacts.
Full Temporal serves as the default lens for answering that question, Taoke broadens the benchmark toward conversion-aware applications, and CasTemp provides a simple anchor method that helps interpret the benchmark without dominating the paper's contribution.

\section{Conclusion}\label{conclusion}

This paper reframes temporal cascade prediction as a benchmarking problem as much as a modeling problem. It delivers a benchmark asset, a protocol suite, and a renewed evaluation standard: Taoke expands the task scope with conversion-aware forecasting, Full Temporal and Inductive User clarify how temporal fidelity and user-transfer leakage should be handled, and the reporting policy distinguishes a default comparison protocol from diagnostic stress tests.
Reliable conclusions in this area require more than another leaderboard result. They require benchmark specifications that make leakage channels visible, richer assets that test more than one target, and reference pipelines that keep trustworthy evaluation practical to run. Our additional Taoke-large stress test reinforces that practicality claim without being conflated with the released benchmark asset itself.

\appendix

\section{Datasets}\label{app:dataset}
We evaluate our method on four real-world datasets spanning social media, academic networks, and e-commerce: three public datasets-\textbf{Twitter}, \textbf{Weibo}, \textbf{APS}—and a newly introduced dataset, \textbf{Taoke}. 
\begin{itemize}
    \item \textbf{Twitter \cite{weng2013viralitytwitter}} contains tweets posted between March 24 and April 25, 2012, along with their retweet cascades. Each cascade represents the diffusion of a specific hashtag.
    
    \item \textbf{Weibo \cite{cao2017deephawkes}} is collected from Sina Weibo, the most popular Chinese microblogging platform, and includes posts published on July 1, 2016, and their subsequent reposts. Each cascade corresponds to the propagation of a single post.
    
    \item \textbf{APS \footnote{\url{https://journals.aps.org/datasets}}} consists of papers published in journals of the American Physical Society (APS) prior to 2017 and their citation relationships. Each cascade models the accumulation of citations for a given paper. Following prior work, we reformulate citation prediction as a cascade forecasting task through appropriate transformation and preprocessing.
    
    \item \textbf{Taoke} is a new real-world dataset introduced in this work. It captures the cascade-like diffusion of product promotions among Taobao promoters (Taoke), where each cascade represents the forwarding of a certain product. Crucially, the dataset also records downstream conversion behavior—i.e., actual consumer purchases—enabling research on second-stage popularity conversion and bridging the gap between cascade prediction and practical applications. It contains all forwarding and transaction records of selected high-sales products on Taobao from August 9, 2025 to August 12, 2025. We also construct an internal extension, \textbf{Taoke-large}, with the same semantics, features, and task definitions but a much larger product pool; it is used only for large-scale efficiency stress testing. The released artifact package is provided in \href{https://github.com/Lucas-PJ/CascadeBench}{our code repository}.
\end{itemize}

\section{Data Privacy and Ethical Considerations}
We confirm that the Taoke dataset has undergone rigorous privacy protection processing to ensure ethical and legal compliance. Specifically, we have implemented the following measures: (1) User Anonymization: All personally identifiable information has been removed and user IDs have been replaced with irreversible pseudonymous tokens. (2) Content Sanitization: The dataset contains no raw text data (e.g., specific production brand) and focuses solely on structured behavioral records (e.g., timestamps, promotions, transactions). (3) Feature Desensitization: Auxiliary features (e.g., product titles) are provided as processed embeddings, preventing reverse-engineering of sensitive semantics. (4) Institutional Review: The dataset and collection methodology were reviewed and approved by the Legal and Security Department of the data source company, confirming no risk of sensitive information leakage. These steps ensure the dataset maintains research utility while strictly preserving user privacy and meeting industrial security standards.

\section{Evaluation and Reporting Notes}
For reproducibility, we note that the artifact label `benchmark\_default' corresponds to \emph{Full Temporal}; the main paper reports popularity results under this protocol, while Random Cascade and Inductive User are used mainly for benchmark diagnosis. The Taoke conversion table is a task-specific subset rather than a duplicate of the full popularity roster. Taoke-large shares the same semantics and task definitions as Taoke, but it is a non-public internal extension used only for large-scale efficiency stress testing, not as a second full benchmark leaderboard. The \href{https://github.com/Lucas-PJ/CascadeBench}{released artifact} exposes the released data, the split scripts, the training/evaluation scripts, and the analysis pipeline for the released benchmark asset.

\section{Baselines}\label{app:Baseline}
We compare our model against a range of established baselines:
\begin{itemize}
    \item \textbf{DeepHawkes}~\cite{cao2017deephawkes}: Models cascades as multiple diffusion paths and employs a GRU network to capture temporal dynamics in diffusion sequences.

    \item \textbf{CasCN}~\cite{chen2019information}: Represents cascades as evolving graph sequences and uses a GNN-LSTM architecture to learn dynamic cascade representations.

    \item \textbf{CasFlow}~\cite{xu2021casflow}: A state-of-the-art approach that learns user representations from both social and cascade graphs, and generates cascade embeddings using a GRU-based sequential model combined with a Variational Autoencoder (VAE).

    \item \textbf{CTCP}~\cite{Lu2023ContinuousTimeGL}: A continuous-time graph learning framework that models inter-cascade dependencies by maintaining dynamic user and cascade representations. It encodes diffusion events as messages and updates node states through an evolutionary learning module with recurrent fusion.

    \item \textbf{CasDo}~\cite{cheng2024information}: Integrates probabilistic diffusion models to capture uncertainty in information spread, by injecting noise during forward propagation and reconstructing cascade embeddings via a reverse denoising process.

    \item \textbf{VNOIP}~\cite{wang2026modeling}: A variational neural ODE-based method that jointly models cascade sequences and popularity-trend trajectories in continuous time. In our benchmark it serves as a strong recent baseline representing trend-aware temporal forecasting.
\end{itemize}

\section{Additional Experimental Results}
For completeness, we provide the full numerical MSLE results corresponding to the ranking matrix in \cref{fig:rank_shift_heatmap}. The detailed performance of all seven baselines under the three evaluation protocols (R: Random Cascade, F: Full Temporal, I: Inductive User) on each dataset is reported in \cref{tab:full_twitter,tab:full_weibo,tab:full_aps,tab:full_taoke} for Twitter, Weibo, APS, and Taoke, respectively. These tables enable direct verification of the rank assignments and facilitate fine-grained comparison beyond the top-3 summary presented in the main text.

\begin{table}[H]
\centering
\caption{Full MSLE results on Twitter across three protocols. Top-3 are color-coded matching Figure~\ref{fig:rank_shift_heatmap}.}
\label{tab:full_twitter}
\setlength{\tabcolsep}{6pt}
\renewcommand{\arraystretch}{1.05}
\begin{tabular}{lccc}
\toprule
\textbf{Model} & \textbf{R} & \textbf{F} & \textbf{I} \\
\midrule
CasTemp    & \cellcolor{blue!50}1.2353 & \cellcolor{blue!50}1.1685 & \cellcolor{blue!50}1.0801 \\
CasCN      & \cellcolor{blue!20}1.2418 & \cellcolor{blue!20}1.1939 & \cellcolor{blue!20}1.0997 \\
DeepHawkes & \cellcolor{blue!8}1.2724  & 1.2702                    & 2.0850 \\
VNOIP      & 1.3280                    & \cellcolor{blue!8}1.2210  & 1.1127 \\
CasFlow    & 1.3035                    & 1.2401                    & 1.1252 \\
CTCP       & 1.5688                    & 1.4474                    & \cellcolor{blue!8}1.1085 \\
CasDo      & 1.9118                    & 2.1014                    & 2.3206 \\
\bottomrule
\end{tabular}
\vskip 0.04in
\end{table}

\begin{table}[H]
\centering
\caption{Full MSLE results on Weibo across three protocols. Top-3 are color-coded matching Figure~\ref{fig:rank_shift_heatmap}.}
\label{tab:full_weibo}
\setlength{\tabcolsep}{6pt}
\renewcommand{\arraystretch}{1.05}
\begin{tabular}{lccc}
\toprule
\textbf{Model} & \textbf{R} & \textbf{F} & \textbf{I} \\
\midrule
CasTemp    & \cellcolor{blue!50}1.5777 & \cellcolor{blue!50}1.4783 & \cellcolor{blue!50}1.5586 \\
CasFlow    & \cellcolor{blue!20}1.6268 & \cellcolor{blue!20}1.6332 & 1.9174 \\
CTCP       & \cellcolor{blue!8}1.7932  & 1.8832                    & \cellcolor{blue!20}1.6359 \\
CasCN      & 1.6669                    & 2.5874                    & 1.7878 \\
DeepHawkes & 1.8345                    & \cellcolor{blue!8}1.7295  & 1.7298 \\
VNOIP      & 2.4950                    & 2.4779                    & 1.8095 \\
CasDo      & 2.9273                    & 2.1425                    & 4.4178 \\
\bottomrule
\end{tabular}
\vskip 0.04in
\end{table}

\begin{table}[H]
\centering
\caption{Full MSLE results on APS across three protocols. Top-3 are color-coded matching Figure~\ref{fig:rank_shift_heatmap}.}
\label{tab:full_aps}
\setlength{\tabcolsep}{6pt}
\renewcommand{\arraystretch}{1.05}
\begin{tabular}{lccc}
\toprule
\textbf{Model} & \textbf{R} & \textbf{F} & \textbf{I} \\
\midrule
CasFlow    & \cellcolor{blue!50}1.0585 & 2.6594                    & 1.8379 \\
CasTemp    & \cellcolor{blue!20}1.0662 & \cellcolor{blue!50}1.9240 & \cellcolor{blue!50}1.3914 \\
CasCN      & \cellcolor{blue!8}1.1924  & \cellcolor{blue!8}2.4956  & 2.6877 \\
VNOIP      & 1.7396                    & \cellcolor{blue!20}2.3991 & \cellcolor{blue!20}1.4960 \\
CTCP       & 1.3559                    & 2.8021                    & \cellcolor{blue!8}1.8317 \\
DeepHawkes & 1.4312                    & 2.5240                    & 2.3433 \\
CasDo      & 2.0807                    & 4.8391                    & 4.7325 \\
\bottomrule
\end{tabular}
\vskip 0.04in
\end{table}

\begin{table}[H]
\centering
\caption{Full MSLE results on Taoke across three protocols. Top-3 are color-coded matching Figure~\ref{fig:rank_shift_heatmap}.}
\label{tab:full_taoke}
\setlength{\tabcolsep}{6pt}
\renewcommand{\arraystretch}{1.05}
\begin{tabular}{lccc}
\toprule
\textbf{Model} & \textbf{R} & \textbf{F} & \textbf{I} \\
\midrule
CasTemp    & \cellcolor{blue!50}1.0113 & \cellcolor{blue!50}0.7984 & \cellcolor{blue!50}0.6715 \\
CasFlow    & \cellcolor{blue!20}2.1328 & \cellcolor{blue!20}2.2877 & \cellcolor{blue!8}1.6461 \\
CasCN      & \cellcolor{blue!8}3.2639  & 3.4317                    & 4.8026 \\
DeepHawkes & 4.3000                    & \cellcolor{blue!8}3.3632  & 2.9352 \\
CTCP       & 6.3728                    & 8.3837                    & \cellcolor{blue!20}1.5410 \\
VNOIP      & 11.8504                   & 6.5742                    & 1.7713 \\
CasDo      & 28.5780                   & 27.1664                   & 3.9857 \\
\bottomrule
\end{tabular}
\end{table}

\newpage

\bibliographystyle{ACM-Reference-Format}
\bibliography{reference}

\end{document}